\documentclass[letterpaper, 10 pt, conference]{ieeeconf}
\IEEEoverridecommandlockouts
\usepackage{cite}
\usepackage{amsmath,amssymb,amsfonts}
\usepackage{algorithmic}
\usepackage{graphicx}
\usepackage{textcomp}
\usepackage{xcolor}
\usepackage{svg}
\usepackage{layouts}
\usepackage{hyperref}
\usepackage{svg}
\usepackage{tikz}
\DeclareMathOperator{\diag}{diag}
\usepackage{siunitx,esvect}

\def\BibTeX{{\rm B\kern-.05em{\sc i\kern-.025em b}\kern-.08em
    T\kern-.1667em\lower.7ex\hbox{E}\kern-.125emX}}
\newcommand{\mb}[1]{\mathbf{#1}}

\author{Sebastien Origer$^{1}$, Christophe De Wagter$^{1}$, Robin Ferede$^{1}$, Guido C.H.E. de Croon$^{1}$,  Dario Izzo$^{2}$
\thanks{*This work was supported by ESA}
\thanks{$^{1}$Micro Air Vehicle Lab of the Faculty
of Aerospace Engineering, Delft University of Technology, 2629 HS
Delft, The Netherlands {\tt\small Sebastien.Origer@outlook.com, C.deWagter@tudelft.nl, R.Ferede@tudelft.nl, G.C.H.E.deCroon@tudelft.nl}}%
\thanks{$^{2}$Advanced Concepts Team, European Space Agency, Keplerlaan 1, 2201 AZ, Noordwijk, The Netherlands. {\tt\small Dario.Izzo@esa.int}}%
}

\title{\LARGE \bf Guidance \& Control Networks for Time-Optimal Quadcopter Flight
}

\begin{document}
\maketitle
\thispagestyle{empty}
\pagestyle{empty}
\begin{abstract}
Reaching fast and autonomous flight requires computationally efficient and robust algorithms. To this end, we train Guidance \& Control Networks to approximate optimal control policies ranging from energy-optimal to time-optimal flight. We show that the policies become more difficult to learn the closer we get to the time-optimal 'bang-bang' control profile. We also assess the importance of knowing the maximum angular rotor velocity of the quadcopter and show that over- or underestimating this limit leads to less robust flight. We propose an algorithm to identify the current maximum angular rotor velocity onboard and a network that adapts its policy based on the identified limit. Finally, we extend previous work on Guidance \& Control Networks by learning to take consecutive waypoints into account. We fly a $4\times3\si{\meter}$ track in similar lap times as the differential-flatness-based minimum snap benchmark controller while benefiting from the flexibility that Guidance \& Control Networks offer.
\end{abstract}
\medskip
\medskip
\noindent Video: \url{https://youtu.be/FrwpODT0HKQ}

\section{Introduction}


\PARstart{M}{icro} air vehicles (MAVs) are very versatile robots with clear applications such as search and rescue missions, entertainment, cinematography, delivery and inspection. Their size, agility, speed, vertical take-off and landing capabilities open up a world of possibilities yet to be explored. In addition, their low cost and the intrinsic challenges of autonomous flight make them the ideal platform to push the frontiers of robotics research. 

\begin{figure}
     \centering
     \vspace{2mm}
    \includegraphics[width=\linewidth]{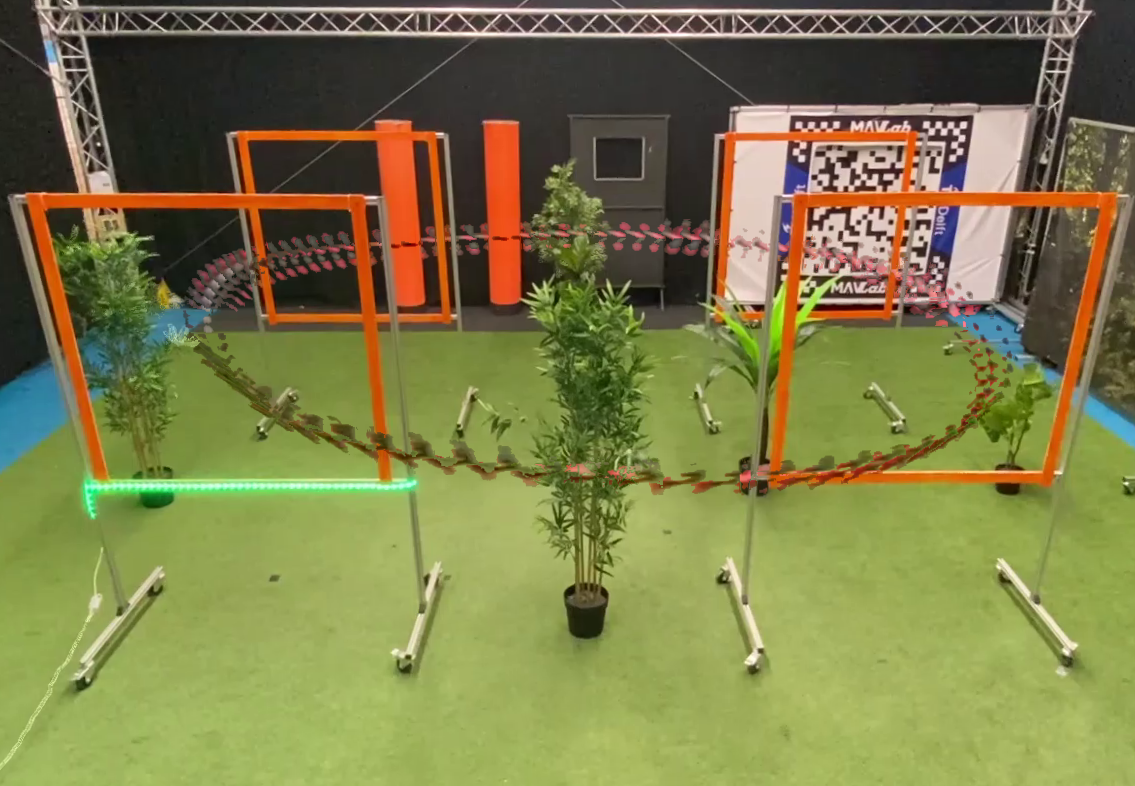}
     \caption{Flight path of the fastest lap using the Parrot Bebop 1 and a Guidance \& Control Network which learned to take two upcoming waypoints into account to compute the optimal control inputs. The track is a $4\times3\si{\meter}$ rectangle, and the four waypoints are positioned at the center of each orange gate. \vspace{-4mm}}
     \label{fig:flight_path}
\end{figure}

Making autonomous and time-optimal flight a reality is relevant as the demand for drones that can perform tasks autonomously is on the rise. Flying time optimally is important as the success of some applications, such as search and rescue missions, hinges on how quickly the drone can reach its destination. In addition, applications that require long flight ranges, such as inspection of offshore wind turbines, also stand to benefit from time-optimal control solutions, as the optimal speed for range is generally relatively fast for multicopters\cite{range_optimal_speed_quads}. This is especially important for drones that are not equipped with a fixed wing, such as quadcopters, as these suffer from limited flight ranges.
Autonomous and time-optimal flight has already been the subject of many studies \cite{alphapilot_win_mavlab,estimation_control_planning,mpccontouring,Mohta_2017,autonomous_drone_race,mellinger_aggressive_man,drone_acrobatics,time_optimal_planning} and drone racing competitions such as the AlphaPilot challenge \cite{alphapilot_win_mavlab} are being organized to further promote research in this field. 
Since aerodynamics effects become more significant for time-optimal flight \cite{neuralmpc} and drones suffer from limited onboard computing power, one of the challenges to overcome is to create robust and computationally efficient control algorithms.

Traditionally, the problem of autonomous flight is broken down into three major steps: perception (state estimation), planning (trajectory generation) and control (trajectory tracking). Some of these steps can be fused together, in fact, one often distinguishes between two types of control strategies:  trajectory tracking methods and trajectory optimization methods. The former first generates the trajectory and then tracks it with a controller, whereas the latter combines both steps together, i.e. computes the control commands directly from the states. 
While state-of-the-art trajectory tracking methods, such as differential-flatness-based-control (DFBC) \cite{MinSnap} or model predictive contouring control (MPCC) \cite{mpccontouring} achieve high speeds, it is usually the first step (the trajectory generation) that is too computationally expensive to be solved onboard of drones \cite{mpccontouring}, unless simplified models are used, such as point-mass models. This is problematic as disturbances are bound to make the quadcopter deviate from the pre-computed trajectory.

Recent advances in machine learning show promising results in the three major steps of autonomous flight \cite{alphapilot_win_mavlab, drone_acrobatics, aggressive_online_control, robin_thesis}. Given a well-posed problem, a large enough training dataset, suitable network architecture and enough computational power, artificial neural nets can today learn to approximate any function up to a certain accuracy. The main machine learning paradigms for guidance and control tasks are reinforcement learning (RL) \cite{drone_acrobatics} and imitation learning \cite{robin_thesis}. RL offers a framework where a network can learn to deal with uncertainty by adding noise to the environment, which is particularly useful when parts of the dynamic system remain unmodelled. However, it is also possible to deal with unmodelled effects using imitation learning \cite{robin_thesis}. In addition, previous work in the context of interplanetary transfers \cite{sanchez_izzo,dario_ekin_earth_venus,IZZO2022} and quadcopters \cite{robin_thesis, aggressive_online_control} has shown that a network can directly learn the optimal state feedback from a large dataset of optimal trajectories. This has led to the term Guidance \& Control Networks (G\&CNETs). G\&CNETs offer a direct mapping from states to raw control commands, they can be inferred at a low computational cost and they are very flexible since there is no need to recompute optimal trajectories.

In this paper, we improve past work on G\&CNETs \cite{robin_thesis} to increase the quadcopter's flight speed, leading to four main contributions. We make the step from energy-optimal to time-optimal control and provide the loss values when training G\&CNETs to approximate the corresponding optimal control policies. We introduce an adaptive scheme that can estimate the maximum angular velocity of the propellers and use it to adjust the commands to remain time-optimal. We develop a training method that allows for the network to output time-optimal raw control commands taking a horizon into account of the next two waypoints. We demonstrate that the two-waypoints network can deal with dynamic waypoint locations during the flight.

We structure the paper as follows. First, the quadcopter model, optimal control problem (OCP), imitation learning procedure and experimental setup are described (Sec.\ref{sec:Methodology}). 
We then consider the task of learning and flying the time-optimal OCP and show how control policies become more difficult to learn as we shift from the energy- to time-optimal control problem (Sec.\ref{sec:Time-optimal quadcopter flight}). 
As one flies more time-optimally, the rotors of the quadcopter saturate more, i.e. the control policy approaches 'bang-bang' control. Given this, we assess the importance of training G\&CNETs at a maximum RPM limit that the rotors of the quadcopter can reach during flight. 
We also propose an algorithm that can identify the current maximum RPM limit in combination with an adaptive G\&CNET which changes its control policy based on the identified limit (\ref{sec:Accounting for the varying max RPM limit of quadcopters}).
Finally, we improve fast quadcopter flight by extending previous research on G\&CNETs from single waypoint to consecutive waypoints flight (Sec.\ref{sec:Consecutive WP flight}). The resulting controller is also compared to another benchmark using time-optimal minimum snap trajectories \cite{MinSnap}.

Fig.\ref{fig:flight_path} shows one lap on a $4\times3\si{\meter}$ track using a G\&CNET which is trained to take two consecutive waypoints into account.

\section{Methodology}\label{sec:Methodology}

\subsection{Quadcopter model}



\begin{figure}[htbp]
  \centering
  \includegraphics[width=\columnwidth]{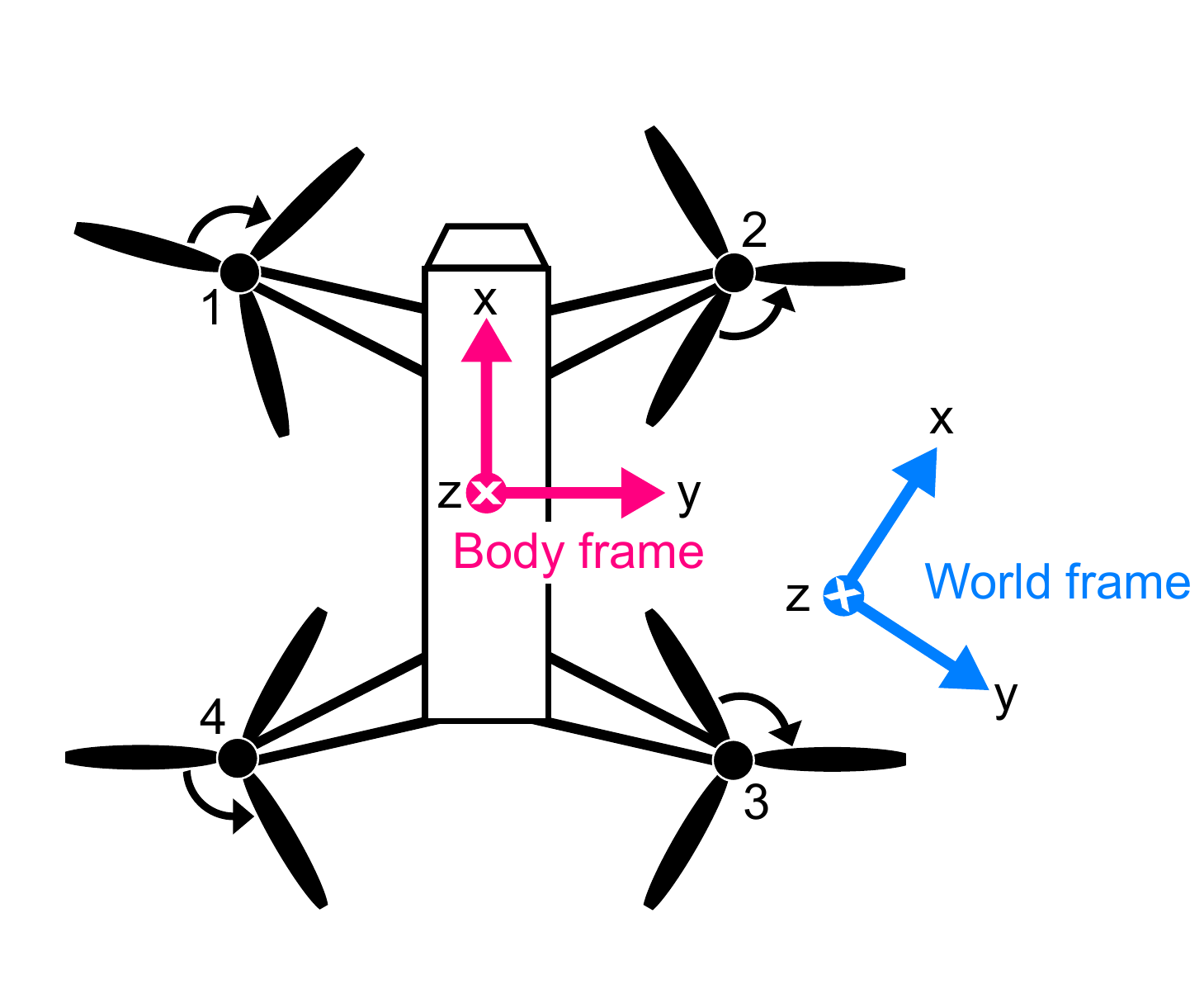}
  \caption{Coordinate frames (Body x-axis points to the front of the drone).\vspace{-5mm}}
  \label{fig:coordframes}
\end{figure}

In this work, we use two coordinate frames as defined in Fig.\ref{fig:coordframes} and a quadcopter model with 19 states and 4 control inputs:

\begin{align*}
    \mb{x} = [\mb{p}, \mb{v}, \mb{\lambda}, \Omega, \mb{\omega},\mb{M}_{ext}]^T \quad \mb{u} = [u_1, u_2, u_3, u_4]^T
\end{align*}

The state vector $\mb{x}$ contains the position $\mb{p}=[x,y,z]$ and velocity $\mb{v}=[v_x,v_y,v_z]$ which are both defined in the world frame. The Euler angles $\mb{\lambda}=[\phi, \theta, \psi]$ which specify the orientation of the body frame, the angular velocities $\mb{\Omega}=[p,q,r]$ in the body frame, the propeller rates $\mb{\omega}=[ \omega_1, \omega_2, \omega_3, \omega_4]$ and external moments disturbances  $\mb{M}_{ext}=[M_{ext,x},M_{ext,y},M_{ext,z}]$ \cite{robin_thesis}. The control inputs $\mb{u} = [u_1, u_2, u_3, u_4]$ are bounded $u_i \in [0,1]$, such that $u_i=0$ and $u_i=1$ correspond to the minimal ($\omega_{min}$) and maximal rotational speed ($\omega_{max}$) of the corresponding propeller, respectively. Specifying the equations of motion (Eq.\ref{eq:dyn}) as:
\begin{equation}
\label{eq:dyn}
f(\mb{x}, \mb{u}) = \left\{ 
\begin{array}{l}
        \dot{\mb{p}} = \mb{v} \\
        \dot{\mb{v}} = \mb{g} + R(\mb{\lambda}) \mb{F} \\
        \dot{\lambda} = Q(\mb{\lambda}) \mb{\Omega} \\
        I \dot{\mb{\Omega}} = - \mb{\Omega} \times I \mb{\Omega} + \mb{M}+ \mb{M}_{ext}\\
        \dot{\omega}  = ((\omega_{max}-\omega_{min}) \mb{u} +\omega_{min} - \omega)/\tau\\
        \dot{\mb{M}}_{ext} = 0
\end{array}
\right.
\end{equation}
where $I=\diag (I_x, I_y, I_z)$ is the moment of inertia matrix and $\mb{g} = [0, 0, g]^T$ with $g=9.81$~\si{\meter\per\second\squared} is the acceleration due to gravity. The rotational matrix $R(\mb{\lambda})$ transforms from the body to the world frame. We use the notation $c_{\theta}$ and $s_{\phi}$ to denote the cosine and sine of the corresponding Euler angle, respectively. 
\begin{equation*}\label{eq:Euler_To World}
R(\mb{\lambda}) = 
\begingroup 
\setlength\arraycolsep{2pt}
\begin{bmatrix}
        c_{\theta}c_{\psi} & -c_{\phi}s_{\psi}+s_{\phi}s_{\theta}c_{\psi} & s_{\phi}s_{\psi}+c_{\phi}s_{\theta}c_{\psi} \\
        c_{\theta}s_{\psi} & c_{\phi}c_{\psi}+s_{\phi}s_{\theta}s_{\psi} & -s_{\phi}c_{\psi}+c_{\phi}s_{\theta}s_{\psi} \\
        -s_{\theta} & s_{\phi}c_{\theta} & c_{\phi}c_{\theta}
\end{bmatrix}
\endgroup
\end{equation*}
and $Q(\mb{\lambda})$ is the inverse transformation matrix: 
\begin{equation*} \label{eq:WorldToEuler}
Q(\mb{\lambda}) = 
\begin{bmatrix}
    1 & \sin{\phi} \tan{\theta}     & \cos{\phi} \tan{\theta} \\
    0 & \cos{\phi}                  & -\sin{\phi} \\
    0 & \sin{\phi} / \cos{\theta}   & \cos{\phi} / \cos{\theta}
\end{bmatrix}
\end{equation*}
The forces $\mb{F} = [F_x, F_y, F_z]^T$ are computed using the thrust and drag model from \cite{thrust_and_drag_model}. Note that the superscript $\square^B$ denotes the body frame, all model parameters are listed in Tab.\ref{tab:model_parameters}.
\begin{align} \label{eq:F_model}
\begin{split}
    F_x &= - k_x v^B_x \sum_{i=1}^4 \omega_i \quad
    F_y = - k_y v^B_y \sum_{i=1}^4 \omega_i\\
    F_z &= -k_\omega \sum_{i=1}^4 \omega_i^2 - k_z v^B_z \sum_{i=1}^4 \omega_i - k_h (v^{B2}_x + v^{B2}_y)
\end{split}
\end{align}
and the moments $\mb{M}=[M_x, M_y, M_z]^T$ are defined as:

\begin{equation} \label{eq:M_model}
\begin{split}
    M_x &= k_p (\omega_1^2 - \omega_2^2 - \omega_3^2 + \omega_4^2) + k_{pv} v^{B}_y\\
    M_y &= k_q (\omega_1^2 + \omega_2^2 - \omega_3^2 - \omega_4^2) + k_{qv} v^{B}_x\\
    M_z &= k_{r1} (-\omega_1 + \omega_2 - \omega_3 + \omega_4) \\
    & + k_{r2} (-\dot{\omega}_1 + \dot{\omega}_2 - \dot{\omega}_3 + \dot{\omega}_4)  - k_{rr} r\\
\end{split}
\end{equation}

We utilize an adaptive method proposed in \cite{robin_thesis}, which accounts for model mismatches in the moment equations (Eq.\ref{eq:M_model}) to make the G\&CNET more robust. The idea is to use domain randomization during the learning process by assuming constant external moment disturbances for each optimal trajectory. The difference between the measured and modelled moments can then be computed onboard and fed to the G\&CNET.

\begin{table*}
    \centering
    \begin{tabular}{c c c c c c c c}
    \hline
    \hline
    $k_x$ & $k_y$ & $k_\omega$ & $k_z$ & $k_h$ & $I_x$ & $I_y$ & $I_z$\\
    \hline
    1.08e-05 & 9.65e-06 & 4.36e-08 & 2.79e-05 & 6.26e-02 &  0.000906 &  0.001242 & 0.002054\\
    \hline
    \hline
    $k_p$ & $k_{pv}$ & $k_q$ & $k_{qv}$ & $k_{r1}$ & $k_{r2}$ & $k_{rr}$  & $\tau$ \\
    \hline
    1.41e-09 & -7.97e-03 & 1.22e-09 & 1.29e-02 & 2.57e-06 & 4.11e-07 & 8.13e-04 & 0.03 \\
    \hline
    \hline
    \end{tabular}
    \caption{Model parameters for the Parrot Bebop 1 quadcopter. The moments of inertia have been taken from \cite{moment_of_inertia}, all other parameters from \cite{robin_thesis}.\vspace{-5mm}}
    \label{tab:model_parameters}
\end{table*}

\subsection{The optimal control problem}

The cost function $J(\mathbf{u}, T)$  (Eq.\ref{eq:cost_function}) minimizes two objectives: the total time of flight $T$ and the energy $\int_{0}^{T} ||\mb{u}(t)||^2 dt$. Both objectives are weighed using the hybridisation parameter $\epsilon$, such that $\epsilon=1$ corresponds to energy-optimal flight and $\epsilon=0$ corresponds to time-optimal flight. While not directly minimizing for time, the energy-optimal term is useful as it creates smooth control inputs which leave more room for errors compared to the fully time-optimal 'bang-bang' control profile.
Denoting $X$ as the state space and $U$ as the set of admissible controls. The optimal control problem considered tries to find the optimal control policy $\mb{u} : [0,1] \rightarrow U$ such that the quadcopter is steered from initial conditions $\mb{x}_0$ to a set of final conditions $S$, while minimizing the cost function $J(\mathbf{u},T)$:

\begin{equation}\label{eq:cost_function}
\begin{split}
    \underset{\mb{u}, T}{\text{minimize}} \quad &J(\mb{u}, T) = (1-\epsilon)T + \epsilon \int_{0}^{T} ||\mb{u}(t)||^2 dt\\
    \text{subject to} \quad &\dot{\mb{x}} = f(\mb{x}, \mb{u})\\
    &\mb{x}(0) = \mb{x}_0 \\
    &\mb{x}(T) \in S
\end{split}    
\end{equation} 

We follow a similar procedure as in \cite{aggressive_online_control} by transcribing the OCP into a Nonlinear Programming (NLP) problem using the Hermite Simpson collocation method. The NLP problem is formulated with the modelling language AMPL \cite{ampl}, discretized into $N+1$ points and solved using a sequential quadratic programming solver called SNOPT \cite{snopt}. AMPL is used as it allows to inform SNOPT on the gradients and Hessian of the problem, making it easier for the solver to converge. We denote the resulting optimal trajectory as $\mb{x}^*_0\ldots\mb{x}^*_N$ with corresponding optimal controls $\mb{u}^*_0\ldots\mb{u}^*_N$.


\subsection{The dataset and learning procedure}

\begin{figure}[t]
\centering
\tikzset{%
  every neuron/.style={
    circle,
    draw,
    minimum size=0.6cm
  },
  neuron missing/.style={
    draw=none, 
    scale=4,
    text height=0.333cm,
    execute at begin node=\color{black}$\vdots$
  },
}
\begin{tikzpicture}[x=1.4cm, y=1cm, >=stealth]

\foreach \m/\l [count=\y] in {1}
  \node [every neuron/.try, neuron \m/.try] (input-\m) at (-1.5,0.25-\y*1.25) {};
\foreach \m [count=\y] in {1,missing,2}
  \node [every neuron/.try, neuron \m/.try ] (hidden-\m) at (-1.0,1.5-\y*1.25) {};
\foreach \m [count=\y] in {1,missing,2}
  \node [every neuron/.try, neuron \m/.try ] (hidden2-\m) at (0.2,1.5-\y*1.25) {};
 \foreach \m [count=\y] in {1,missing,2}
  \node [every neuron/.try, neuron \m/.try ] (hidden3-\m) at (1.4,1.5-\y*1.25) {};
\foreach \m [count=\y] in {1,2,3,4}
  \node [every neuron/.try, neuron \m/.try ] (output-\m) at (2.6,1.5-\y*1.0) {};

\foreach \l [count=\i] in {\mb{x}}
  \draw [<-] (input-\i) -- ++(-0.75,0)
    node [above, midway] {$\l$};
\foreach \l [count=\i] in {1,2,3,4}
  \draw [->] (output-\i) -- ++(0.75,0)
    node [above, midway] {$u_\l$};

\foreach \i in {1}
  \foreach \j in {1,...,2}
    \draw [->] (input-\i) -- (hidden-\j);
\foreach \i in {1,...,2}
  \foreach \j in {1,...,2}
    \draw [->] (hidden-\i) -- (hidden2-\j);
\foreach \i in {1,...,2}
  \foreach \j in {1,...,2}
    \draw [->] (hidden2-\i) -- (hidden3-\j);
\foreach \i in {1,...,2}
  \foreach \j in {1,2,3,4}
    \draw [->] (hidden3-\i) -- (output-\j);

\foreach \l [count=\x from 1] in {(ReLU)\\120\\neurons, (ReLU)\\120\\neurons,(ReLU)\\120\\neurons, (Sigmoid) \\ Output\\layer}
  \node [align=center, above] at (-2.2+\x*1.2,1) {\l};
\end{tikzpicture}
\caption{Feed forward network architecture.\vspace{-5mm}}
\label{fig:ffnn}
\end{figure}
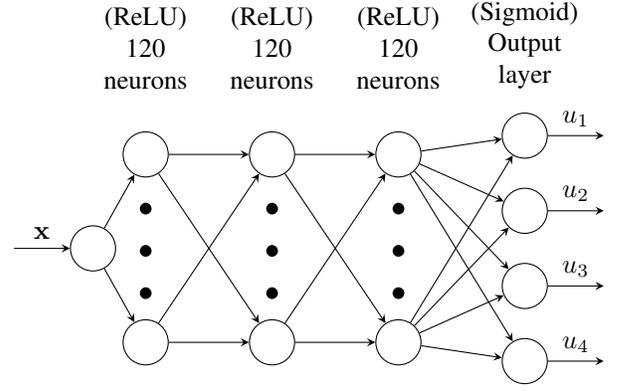

The G\&CNETs in this work are trained via imitation / supervised learning using datasets of optimal state-action pairs, where the states serve as features and the controls as labels. Hence each entry in the dataset is $(\mb{x}_i^*, \mb{u}_i^*) \quad i = 0, \ldots, N$. Learning the optimal state feedback policy $\text{G\&CNET}(\mb{x}_i^*) \approx \mb{u}_i^*$ is possible because of the existence and uniqueness of an optimal state-feedback which is a result of the Hamilton-Jacobi Bellman equations \cite{sanchez_izzo}. The general network architecture used throughout this work is a feed-forward neural net as depicted in Fig.\ref{fig:ffnn}.
We use the mean squared error as loss function:
\begin{equation*}
    \mathcal L=|| \text{G\&CNET}(\mb{x}_i^*) - \mb{u}_i^* ||^2
\end{equation*}

The dataset is split into training data ($80\%$) and validation data ($20\%$). The weights of the network are updated using the Adam optimizer \cite{kingma2014adam} without weight decay. The starting learning rate is set to $l=0.1\cdot 10^{-2}$ and a scheduler is used to reduce the learning rate by a factor of $f=0.9$ whenever the loss on the validation dataset plateaus for $p=6$ epochs. In order to facilitate the learning process the features are normalized and an addition output layer is added at the end to map the control inputs $u_i \in [0,1]$ to the corresponding RPMs.

\subsection{Experimental setup}

The experimental platform in this work is the Parrot Bebop 1 quadcopter in combination with the open-source Paparazzi UAC software \cite{paparazzi}. This drone has been designed for the selfie-drone market, hence it is not a racing drone. Having a maximum thrust-to-weight ratio of $\sim1.7$ it is not able to reach the same speeds as drones in the autonomous drone racing literature. However, this also means that it will allow us to reach saturation of its controls in the relatively small flight space at our disposal and that it is safer to test with. These latter properties were the main motivation for selecting the Parrot Bebop 1 for this study. The Bebop is equipped with an MPU6050 IMU and a Parrot P7 dual-core Cortex A9 CPU which we use to run our code in real-time onboard the drone. In addition, the quadcopter can measure the angular velocities of the four propellers. We fly the quadcopter at the faculty of Aerospace Engineering (TU Delft) inside The Cyberzoo, which is a 10-by-10 meter laboratory equipped with a motion-capture system (Optitrack). We send position, velocity and attitude measurements to the drone in real-time and fuse these with the IMU measurements using an Extended Kalman filter to provide accurate state estimation to the G\&CNET. The G\&CNET then directly outputs RPM commands which are sent to the motors. In the case of the differential-flatness-based-controller we use an INDI controller to track the reference trajectory.

\section{Time-optimal quadcopter flight}\label{sec:Time-optimal quadcopter flight}

In this section, we train G\&CNETs to approximate different control policies, from fully energy-optimal ($\epsilon=1.0$) to fully time-optimal ($\epsilon=0.0$). We simulate the response of the G\&CNET with $\dot{\mb{x}} = f(\mb{x}, \text{G\&CNET}(\mb{x}))$ and show that for close to time-optimal flight, reaching a low loss $\mathcal{L}$ not only becomes more difficult but it is also more critical as the quadcopter has less control authority to recover from errors compared to energy-optimal flight.

\subsection*{The datasets and network architecture}

\begin{figure}[htbp]
  \centering
  \includegraphics[width=\columnwidth]{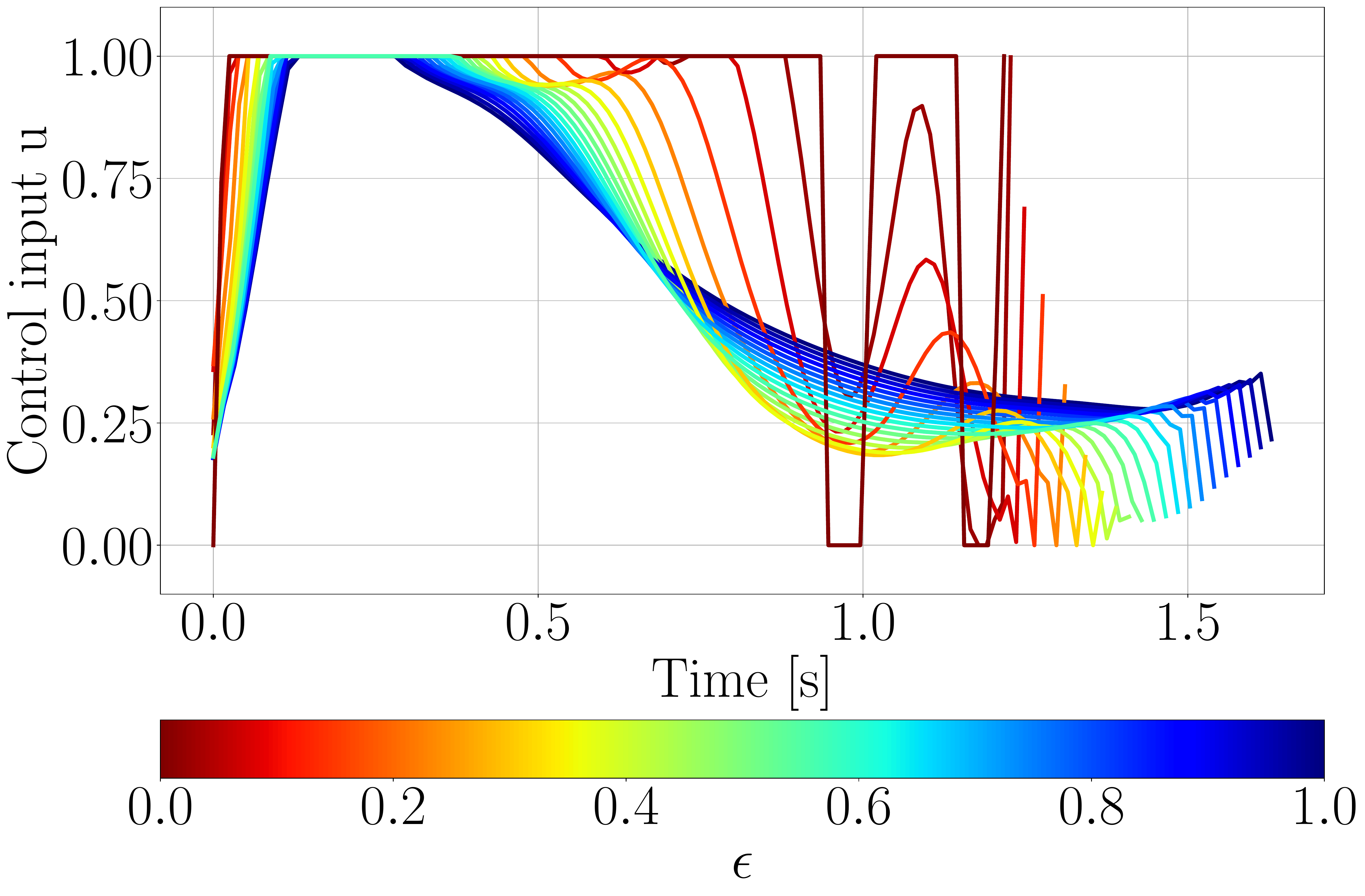}
  \caption{Optimal control input $u^*$ for different $\epsilon$ (only one rotor is shown). All trajectories have the same initial and final conditions.\vspace{-2mm}}
  \label{fig:control_inp2}
\end{figure}

We generate all training datasets by uniformly sampling the initial conditions for each trajectory in the following bounds (where the target waypoint is the reference, see the top figure of Fig.\ref{fig:opt_traj_training} in App.\ref{app:opt_traj_training} as an example):

\begin{align*}
    x_0 &\in [-5.0,-2.0],\si{\meter} &
    y_0 &\in [-1.0,1.0],\si{\meter} \\
    z_0 &\in [-0.5,0.5],\si{\meter} &
    v_{x_0}&\in [-0.5,5.0],\si{\meter\per\second} \\
    v_{y_0}&\in [-3.0,3.0],\si{\meter\per\second} &
    v_{z_0}&\in [-1.0,1.0],\si{\meter\per\second} \\
    \phi_0 &\in [-40,40],\text{deg} &
    \theta_0 &\in [-40,40],\text{deg} \\
    \psi_0 &\in [-60,60],\text{deg} &
    p_0 &\in [-1,1],\si{\radian\per\second} \\
    q_0 &\in [-1,1],\si{\radian\per\second} &
    r_0 &\in [-1,1],\si{\radian\per\second}
\end{align*}
The RPM limits are set to $\omega_{min} = 3000$ and $\omega_{max}=12000$, respectively, and the initial rotational speeds are sampled in:
\begin{align*}
    \omega_{i_0} \in [-\omega_{min},\omega_{max}],\text{RPM} \quad i=1,\ldots,4
\end{align*}
For each trajectory we assume constant external moment disturbances sampled in these bounds:
\begin{align*}
 M_{ext,x} &\in [-0.04,0.04],\si{\newton\meter} \\ M_{ext,y} &\in [-0.04,0.04],\si{\newton\meter} \\ M_{ext,z} &\in [-0.01,0.01],\si{\newton\meter} 
\end{align*}
We set the desired final states for each trajectory to: $\mb{p}_f = \mb{0}$ $\si{\meter}$, $\psi_f = 45^\circ$, $\mb{\Omega}_f = \mb{0}$ $\si{\radian\per\second}$ and $\dot{\mb{\Omega}}_f = \mb{0}$ $\si{\radian\per\second\squared}$. The final velocity $\mb{v}_f$ is constrained such that its direction coincides with the final heading $\psi_f = 45^\circ$, the final velocity magnitude and all remaining states are left free. These specific constraints are chosen as they allow the G\&CNET to fly a variety of tracks containing only turns to the right by moving the target waypoint right before the quadcopter reaches it. Sec.\ref{sec:Consecutive WP flight} goes into more detail regarding single and consecutive waypoints flight. Each dataset contains 10,000 optimal trajectories which are all sampled in $N=199$ points. We choose to train each individual G\&CNET for $p=10$ epochs with a training batch size of 256.
Fig.~\ref{fig:control_inp2} shows how a typical optimal control input $u^*$ for one of the four rotors varies for different $\epsilon$. Given the same initial and final conditions, $u^*$ varies from a smooth control profile ($\epsilon=1.0$) to a so-called 'bang-bang' control profile ($\epsilon=0.0$) for fully time-optimal flight, where at all times at least one of the four rotors is saturating. Note that the time-optimal solution completes this trajectory in $1.2\si{\second}$ compared to $1.65\si{\second}$ for the energy-optimal case.

\subsection*{Results \& Discussion}

We report the resulting loss $\mathcal{L}$ and corresponding mean control error $[\%]$ in Tab.~\ref{tab:mseerrors}. Clearly, as we weigh the time-optimal objective in Eq.~\ref{eq:cost_function} more heavily by decreasing $\epsilon$, the loss $\mathcal{L}$ goes up. The reason the time-optimal policy is more difficult to learn is the high number of switching times as depicted in Fig.~\ref{fig:control_inp2}. The large gradients of the resulting topology are more difficult to approximate precisely than the smooth continuous control profile for energy-optimal flight. 

In practice, this issue can be mitigated by increasing the training dataset size, training for longer and eventually increasing the size of the network. Though one must mention that a larger network will directly impact the frequency at which the network can be inferred onboard the drone. In addition, for time-optimal flight, bringing the loss $\mathcal{L}$ down will not solve all problems. Time-optimal flight means the drone operates at the edge of its flight envelope, leaving no room for control authority. This means that recovering from small control errors becomes much more difficult than for energy-optimal flight.

We test all G\&CNETs in simulation by numerically integrating $\dot{\mb{x}} = f(\mb{x}, \text{G\&CNET}(\mb{x}))$ using Scipy (explicit Runge-Kutta integration method of order $5$ \cite{rungekutta}). No perturbations are added to these simulations and the same equations of motion and actuator delay as the ones used to solve the optimal trajectories are used. The solver chooses the step size and no zero-order hold is implemented. Even with these conditions, it becomes hard to maintain stable flight for $\epsilon=0.1$ and $\epsilon=0.0$. This suggests that for close to time-optimal flight, a training loss of $\mathcal{L}=1.12\cdot 10^{-3}$ (mean control error of $\pm3.35\%$) is an upper bound to maintain stability.

\begin{table}[htbp]
    \centering
    \begin{tabular}{c|cc}
$\epsilon$ &  Loss $\mathcal{L}$ & Control error [$\%$] \\\hline \hline

1.0 &  $1.24\cdot 10^{-4}$  &  $\pm1.12$  \\
 \hline

0.95 &  $1.41\cdot 10^{-4}$  &  $\pm1.19$  \\
 \hline
 
0.9 &  $1.38\cdot 10^{-4}$  &  $\pm1.18$  \\
 \hline

0.85 &  $1.35\cdot 10^{-4}$  &  $\pm1.16$  \\
 \hline

0.8 &  $1.46\cdot 10^{-4}$  &  $\pm1.21$  \\
 \hline

0.75 &  $1.48\cdot 10^{-4}$  &  $\pm1.22$  \\
 \hline

0.7 &  $1.60\cdot 10^{-4}$  &  $\pm1.27$  \\
 \hline

0.65 &  $1.72\cdot 10^{-4}$  &  $\pm1.31$  \\
 \hline

0.6 &  $1.97\cdot 10^{-4}$  &  $\pm1.40$  \\
 \hline

0.55 &  $1.79\cdot 10^{-4}$  &  $\pm1.34$  \\
 \hline

0.5&  $2.05\cdot 10^{-4}$  &  $\pm1.43$  \\
 \hline

0.45&  $2.37\cdot 10^{-4}$  &  $\pm1.54$  \\
 \hline

0.4&  $2.99\cdot 10^{-4}$  &  $\pm1.73$  \\
 \hline

0.35&  $2.96\cdot 10^{-4}$  &  $\pm1.72$  \\
 \hline

0.3&  $3.39\cdot 10^{-4}$  &  $\pm1.84$  \\
 \hline

0.25&  $4.82\cdot 10^{-4}$  &  $\pm2.20$  \\
 \hline

0.2&  $5.25\cdot 10^{-4}$  &  $\pm2.29$  \\
 \hline

0.15&  $7.08\cdot 10^{-4}$  &  $\pm2.66$  \\
 \hline

0.1 &  $1.12\cdot 10^{-3}$  &  $\pm3.35$  \\
 \hline

0 &  $7.01\cdot 10^{-3}$  &  $\pm8.37$  \\
 \hline 

    \end{tabular}
\caption{Mean squared errors on validation datasets and corresponding control errors for G\&CNETs trained on different cost functions.\vspace{-0mm}} \label{tab:mseerrors}
\end{table}

\section{Accounting for the varying maximum angular velocity of propellers}\label{sec:Accounting for the varying max RPM limit of quadcopters}

\begin{figure}[htbp]
  \centering
  \includegraphics[width=\columnwidth]{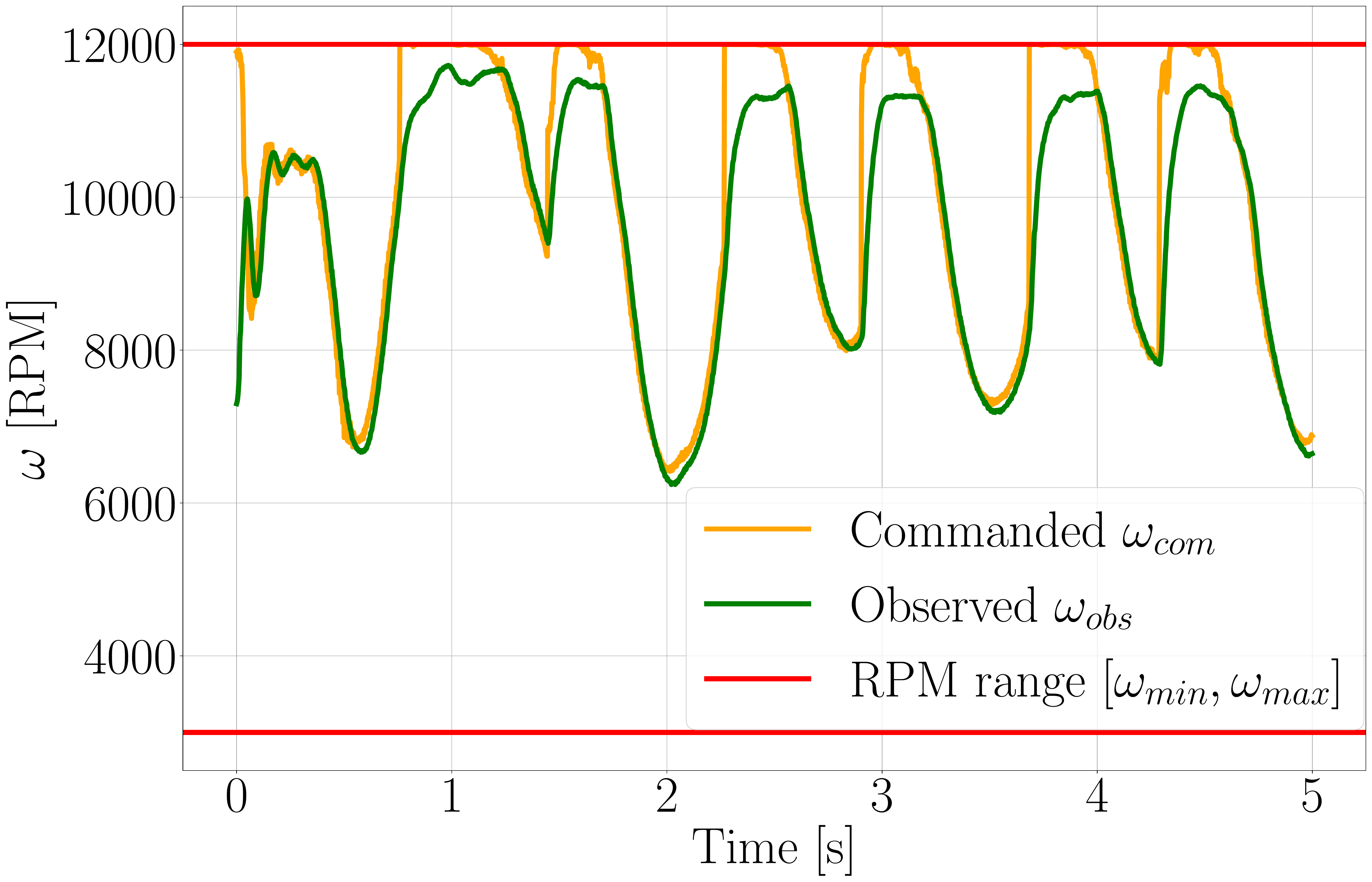}
  \caption{Discrepancy between commanded and observed angular velocity $\omega$ during a real flight with $\epsilon=0.35$ (only one rotor is shown).\vspace{-2mm}}
  \label{fig:rpm_comm_vs_obs}
\end{figure}

In this section, we investigate the importance of flying at a reachable maximum RPM limit $\omega_{max}$ by looking at how under- and overshooting this limit affects the robustness of the flight. The maximum angular velocity of rotors can either be wrongly identified in the first place, or momentarily change during flight due to varying aerodynamic load on the propellers or even decrease over time as the battery drains out. We have observed a steady drop of $\dot{\omega}_{max} = -1 \text{ RPM}\si{\per\second}$ on the Parrot Bebop 1 over a test flight of $6\si{\minute}$ (See Appendix \ref{app:battery_effect}). Time-optimal flight is characterised by control profiles that saturate the rotors for a considerable portion of time, consider Fig.~\ref{fig:rpm_comm_vs_obs} which shows the commanded and observed angular velocities of one rotor during a real flight with $\epsilon=0.35$. The upper RPM limit $\omega_{max}=12000$ cannot be reached. This begs the question, how crucial is it to fly at the correct $\omega_{max}$? 

\begin{figure}[htbp]
  \centering
  \includegraphics[width=\columnwidth]{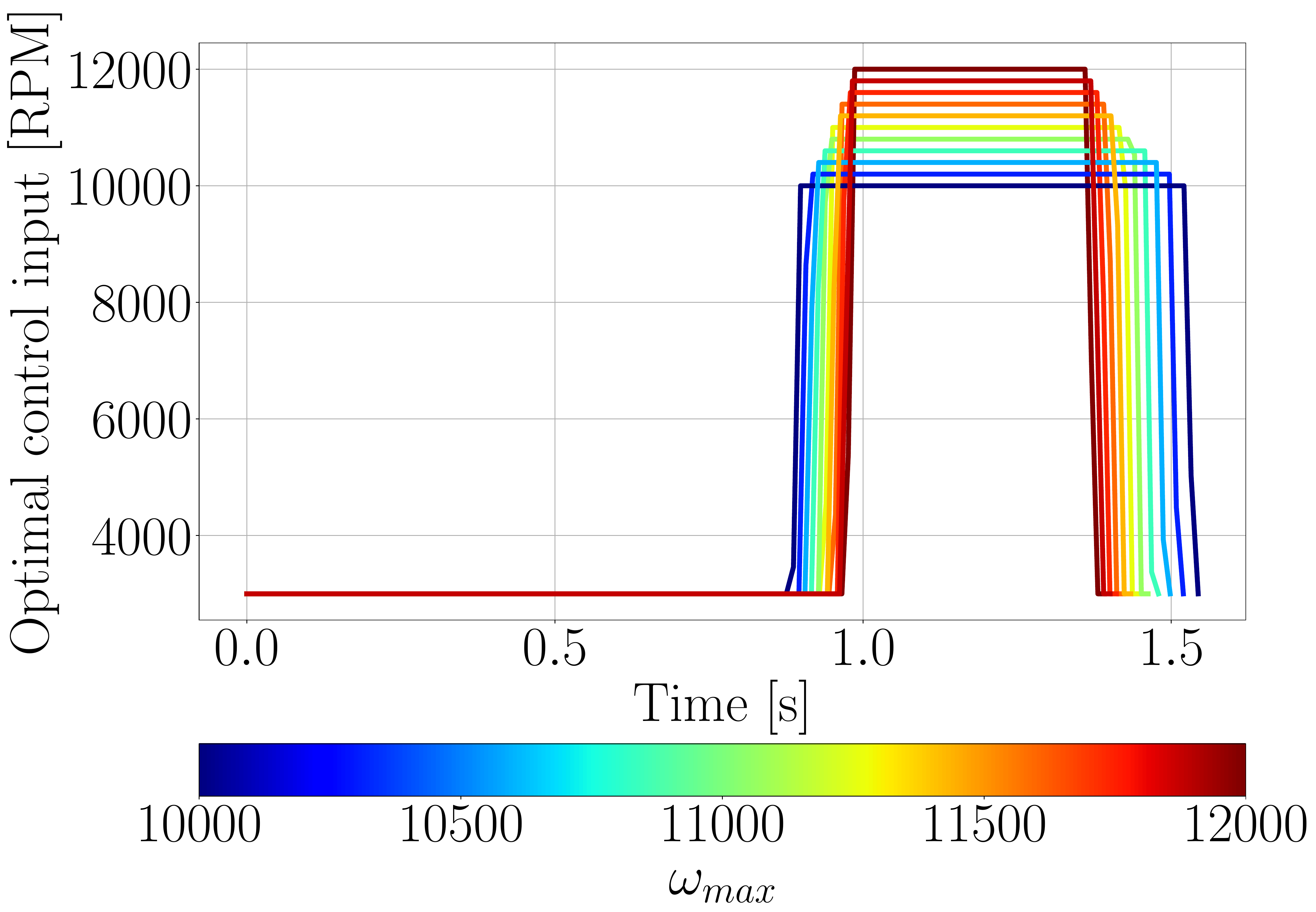}
  \caption{Optimal control input [RPM] for different $\omega_{max}$ (only one rotor is shown). All trajectories have the same initial and final conditions. The goal is to perform a time-optimal ($\epsilon=0.0$) landing from a height of $5\si{\meter}$. Given the symmetry of this OCP all rotors receive the same control input.\vspace{-5mm}}
  \label{fig:control_input_diff_max_rpm}
\end{figure}

Let's consider a range of $\omega_{max} [10000,12000]$ RPM. Fig.~\ref{fig:control_input_diff_max_rpm} shows how the optimal control solution differs for different $\omega_{max}$ in the case of a time-optimal ($\epsilon=0.0$) landing. The quadcopter starts from hover at a height of $5\si{\meter}$ and needs to reach the following final conditions: $\mb{p}_f = \mb{0}$, $\mb{v}_f = \mb{0}$, attitude $\mb{\lambda}_f = \mb{0}$ and angular rates $\mb{\Omega}_f = \mb{0}$. No external moment disturbances $\mb{M}_{ext}$ are applied in this case. Given the symmetry of this OCP, the optimal control solution for all four rotors is the same. In the case where $\omega_{max}=12000$ the rotors start saturating $0.1\si{\second}$ later than in the case where $\omega_{max}=10000$. Since the optimal control solutions here consist in applying the maximal breaking force from a certain switching point onwards until the end, overestimating $\omega_{max}$ will always result in a crash since the quadcopter will start breaking too late. This case also suggests that having the observed angular velocities of the propellers $\mb{\omega}$ as state feedback is not sufficient to mitigate the effects of incorrectly identifying $\omega_{max}$ because when the drone observes that the rotors do not reach the desired $\omega_{max}$, it is already too late. Finally, precise switching times between minimal and maximal control inputs are required to maintain stable flight, especially because little to no control authority is left as $\epsilon$ approaches zero.

\subsection*{Peak tracker algorithm}

\begin{figure}[htbp]
  \centering
  \includegraphics[width=\columnwidth]{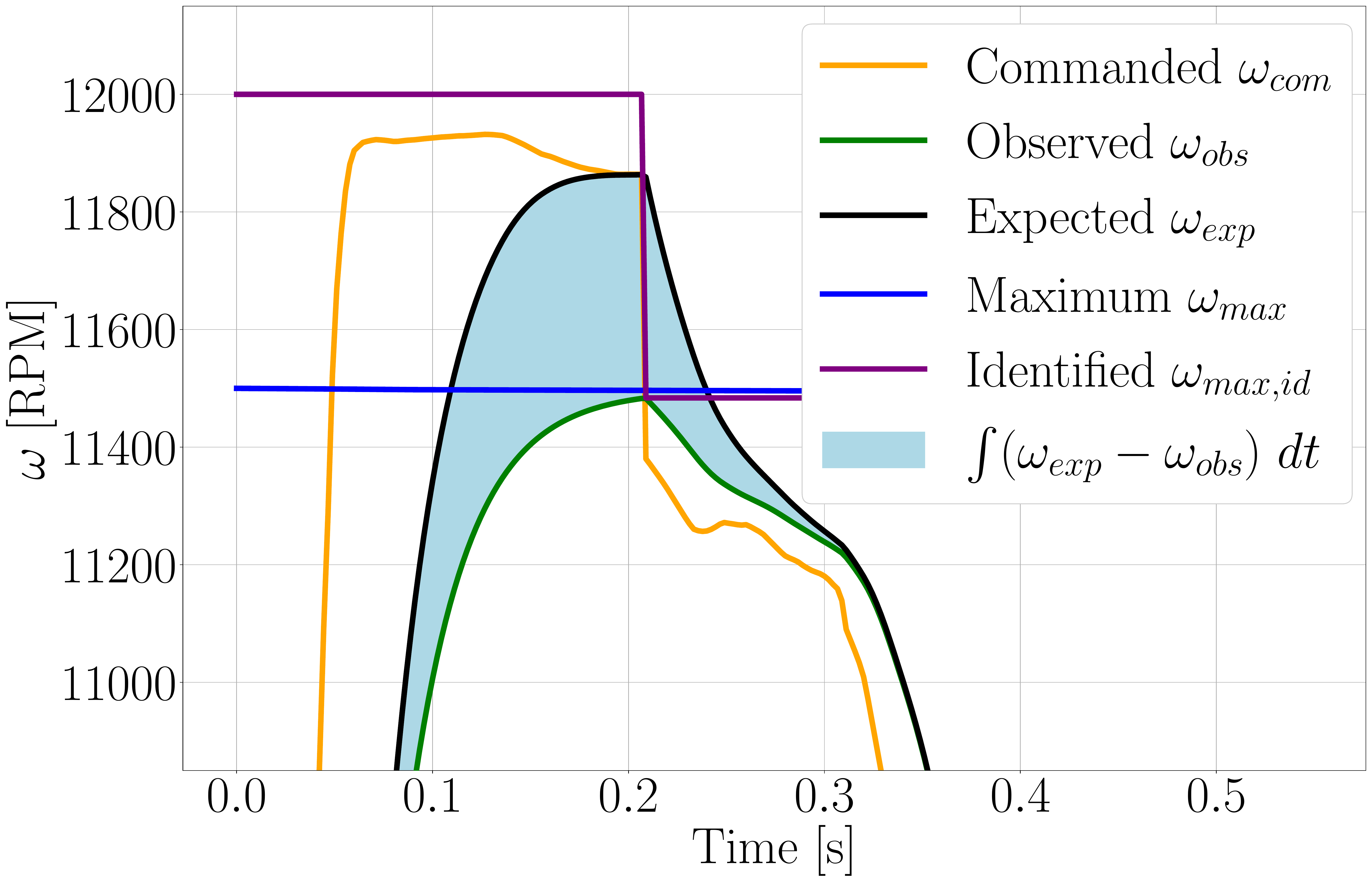}
  \caption{Identifying the current $\omega_{max}$ (blue line) with the peak tracker algorithm.\vspace{-5mm}}
  \label{fig:peak_tracker}
\end{figure}

In the case where the initial guess for $\omega_{max}$ is set too high, the new limit can be identified onboard as soon as it is observable. We propose a peak tracker algorithm that sets $\omega_{max}$ to the highest observed $\omega_{obs}$ whenever the integral $\int^t_{t-\Delta t}(\omega_{exp}-\omega_{obs})dt$ surpasses a certain threshold $\text{p}_{thresh}$, see $t=0.21\si{\second}$ in Fig.~\ref{fig:peak_tracker}. The expected $\omega_{exp}$ can be computed onboard the drone by taking the first order delay of the commanded $\omega_{com}$ (Eq.\ref{eq:dyn}). Depending on the RPM range $[\omega_{min},\omega_{max}]$ the parameters of this algorithm to be tuned are: the time window $\Delta t$ over which the integral is computed and the last peak in observed $\omega_{obs}$ is recorded and the threshold $\text{p}_{thresh}$ which triggers a change in $\omega_{max}$. One should note that another obvious approach would be to model $\omega_{max}$ as a function of the quadcopter states $\mb{x}$ and its battery voltage. 
However, this would require system identification every time one changes the drone. The advantage of our algorithm is that it can easily be used for any drone, so long as the angular velocity of the propellers can be measured onboard. In addition, our algorithm could account for lower $\omega_{max}$ due to any unexpected failure that cannot be modelled.
The current setup can only correct $\omega_{max}$ after overshooting it, otherwise, the limit is not observable.
We tested the RPM peak tracker algorithm both in simulation and on the real quadcopter. In both cases, when the initial $\omega_{max}$ is too high, it takes less than $\Delta t=0.13\si{\second}$ after one of the rotors saturates to correct the limit. We also simulated whether the peak tracker can continuously correct for a slowly decreasing $\omega_{max}$ (due to for instance the decrease in battery voltage). The peak tracker can keep the error between the identified limit and the correct $\omega_{max}$ below $70\text{ RPM}$ at all times.

\subsection*{Adaptive G\&CNET}

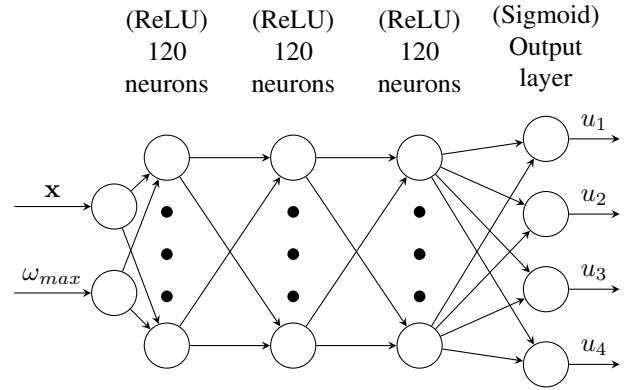
\begin{figure}[htbp]
\centering
\tikzset{%
  every neuron/.style={
    circle,
    draw,
    minimum size=0.6cm
  },
  neuron missing/.style={
    draw=none, 
    scale=4,
    text height=0.333cm,
    execute at begin node=\color{black}$\vdots$
  },
}
\begin{tikzpicture}[x=1.4cm, y=1cm, >=stealth]

\foreach \m/\l [count=\y] in {1,2}
  \node [every neuron/.try, neuron \m/.try] (input-\m) at (-1.5,0.75-\y*1.15) {};
\foreach \m [count=\y] in {1,missing,2}
  \node [every neuron/.try, neuron \m/.try ] (hidden-\m) at (-1.0,1.5-\y*1.25) {};
\foreach \m [count=\y] in {1,missing,2}
  \node [every neuron/.try, neuron \m/.try ] (hidden2-\m) at (0.2,1.5-\y*1.25) {};
 \foreach \m [count=\y] in {1,missing,2}
  \node [every neuron/.try, neuron \m/.try ] (hidden3-\m) at (1.4,1.5-\y*1.25) {};
\foreach \m [count=\y] in {1,2,3,4}
  \node [every neuron/.try, neuron \m/.try ] (output-\m) at (2.6,1.5-\y*1.0) {};

\foreach \l [count=\i] in {\mb{x}, \omega_{max}}
  \draw [<-] (input-\i) -- ++(-0.95,0)
    node [above, midway] {$\l$};
\foreach \l [count=\i] in {1,2,3,4}
  \draw [->] (output-\i) -- ++(0.7,0)
    node [above, midway] {$u_\l$};

\foreach \i in {1,2}
  \foreach \j in {1,...,2}
    \draw [->] (input-\i) -- (hidden-\j);
\foreach \i in {1,...,2}
  \foreach \j in {1,...,2}
    \draw [->] (hidden-\i) -- (hidden2-\j);
\foreach \i in {1,...,2}
  \foreach \j in {1,...,2}
    \draw [->] (hidden2-\i) -- (hidden3-\j);
\foreach \i in {1,...,2}
  \foreach \j in {1,2,3,4}
    \draw [->] (hidden3-\i) -- (output-\j);

\foreach \l [count=\x from 1] in {(ReLU)\\120\\neurons, (ReLU)\\120\\neurons,(ReLU)\\120\\neurons, (Sigmoid) \\ Output\\layer}
  \node [align=center, above] at (-2.2+\x*1.2,1) {\l};
\end{tikzpicture}
\caption{Feed forward network architecture for an adaptive G\&CNET to changes in $\omega_{max}$.\vspace{-5mm}}
\label{fig:ffnn_adaptive_om_max}
\end{figure}

We considered training a G\&CNET on a range of optimal trajectories with different $\omega_{max}$. Instead of only learning the mapping between states $\mb{x}$ and controls $\mb{u}$, we add $\omega_{max}$ as one of the features. This allows the G\&CNET to adapt during flight by using the output of the peak tracker algorithm as an additional input, see the new architecture in Fig.~\ref{fig:ffnn_adaptive_om_max}.

\subsection*{Results and discussion}

\begin{figure}[htbp]
  \centering
  \includegraphics[width=\columnwidth]{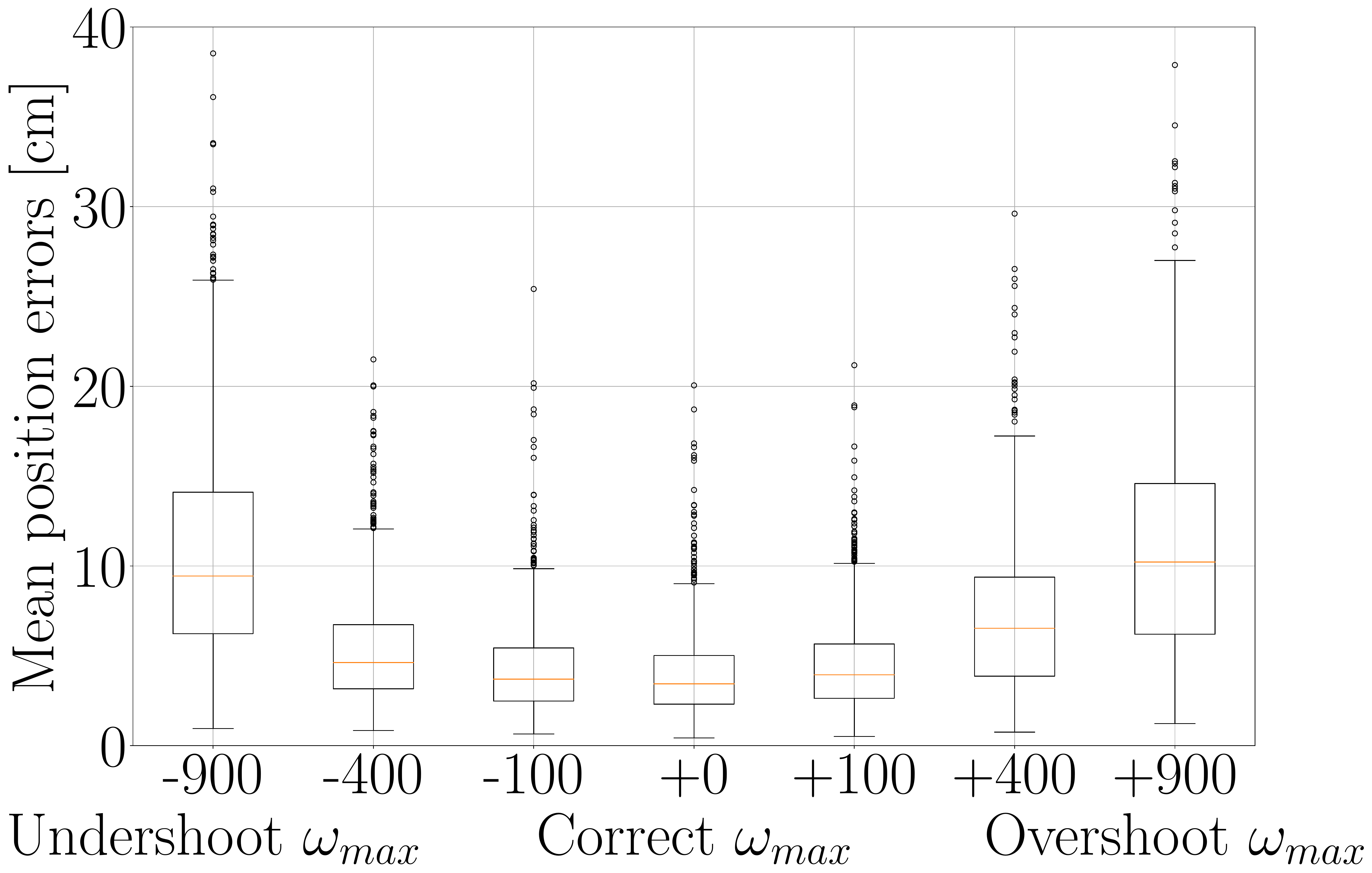}
  \caption{Mean position errors [cm] from 10,000 optimal trajectories (1000 nodes per trajectory). For each boxplot the G\&CNET either undershoots, overshoots or is exactly at the correct $\omega_{max}\text{ [RPM]}$.\vspace{-0mm}}
  \label{fig:box_plots_mean_pos_errors}
\end{figure}

We first evaluate the effect of over- and undershooting $\omega_{max}$ in simulation. We train two adaptive G\&CNETs using the same bounds for the initial and final conditions as in Sec.\ref{sec:Time-optimal quadcopter flight} except that each training dataset now contains 100,000 optimal trajectories and $\epsilon=0.4$. The first G\&CNET is trained on optimal trajectories where $\omega_{max}$ is sampled uniformly in $[10000,11000]$ RPM and the second G\&CNET $[11000,12000]$ RPM. Note that for this analysis, one could just as well train multiple G\&CNETs, each specialized for only one value for $\omega_{max}$. The adaptive G\&CNETs make it easier to quickly see how changing $\omega_{max}$ affects its performance. 
We generate an evaluation dataset separately which contains 10,000 optimal trajectories, all of which use $\omega_{max}=11100$ RPM. This $\omega_{max}$ value was chosen in order to simulate over a large range for over- and undershooting this limit while staying close to the real limit of the Parrot Bebop 1. Given that the optimal trajectories are not the true analytical optimal solutions but are solved with a direct method, one can expect considerable numerical noise due to integration errors between the nodes. To alleviate some of this noise, we augment all 10,000 optimal trajectories using a node-doubling technique. The OCPs considered here are too complex for SNOPT to converge for $N>400$ nodes if no good initial guess is provided. Hence we solve the OCP for $N=100$, then interpolate the solution $\mb{x}^*$ and $\mb{u}^*$ using quadratic splines and finally project the interpolant on a new grid of nodes (e.g. $N=200$) to have a good initial guess for the solver. Repeating this process allowed us to generate a high-fidelity evaluation dataset of 10,000 trajectories, with $N=1000$ nodes each (which translates to $5\si{\milli\meter}$ between two consecutive nodes for a $5\si{\meter}$ long optimal trajectory). Fig.~\ref{fig:box_plots_mean_pos_errors} shows the mean position errors [cm] from these 10,000 optimal trajectories when simulating the response of the G\&CNETs $\dot{\mb{x}} = f(\mb{x}, \text{G\&CNET}(\mb{x}))$ starting from the same initial conditions as the trajectories in the evaluation dataset. By manually changing the input $\omega_{max}$ to the network, we can simulate how this affects the deviation from the optimal trajectories. Even when the G\&CNET knows the correct limit (boxplot in the center of Fig.~\ref{fig:box_plots_mean_pos_errors}) its mean position error is around $4\si{\centi\meter}$, which is due to the nonzero loss during training. We see a similar trend for over- and undershooting $\omega_{max}$, the larger the difference, the larger the mean position error. This suggests that incorrectly identifying $\omega_{max}$ impacts the robustness of the flight.

However, contrary to our expectations, even large deviations from $\omega_{max}$ are not the most critical contributor to the reality gap. In our experiments in simulation and on the real quadcopter, we were able to fly even when $\omega_{max}$ was off by $+1000\text{ RPM}$. Nevertheless, as we approach time-optimal flight, there is less room for error and flying as closely as possible to the optimal trajectory is relevant for robustness.

\begin{figure*}[htbp]
  \centering
  \begin{minipage}[b]{0.32\textwidth}
    \includegraphics[width=\textwidth]{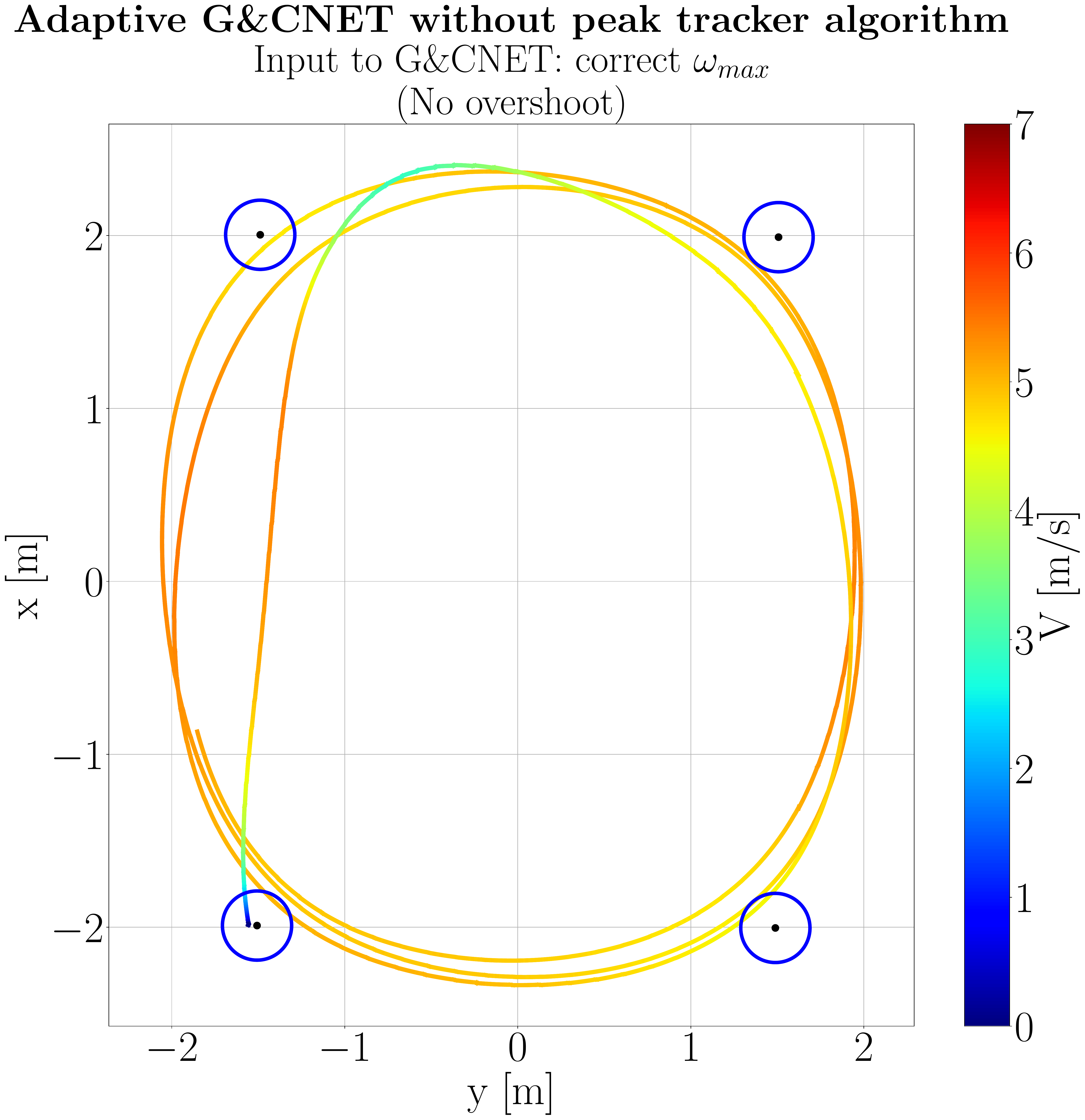}
  \end{minipage}
  \hfill
  \begin{minipage}[b]{0.32\textwidth}
    \includegraphics[width=\textwidth]{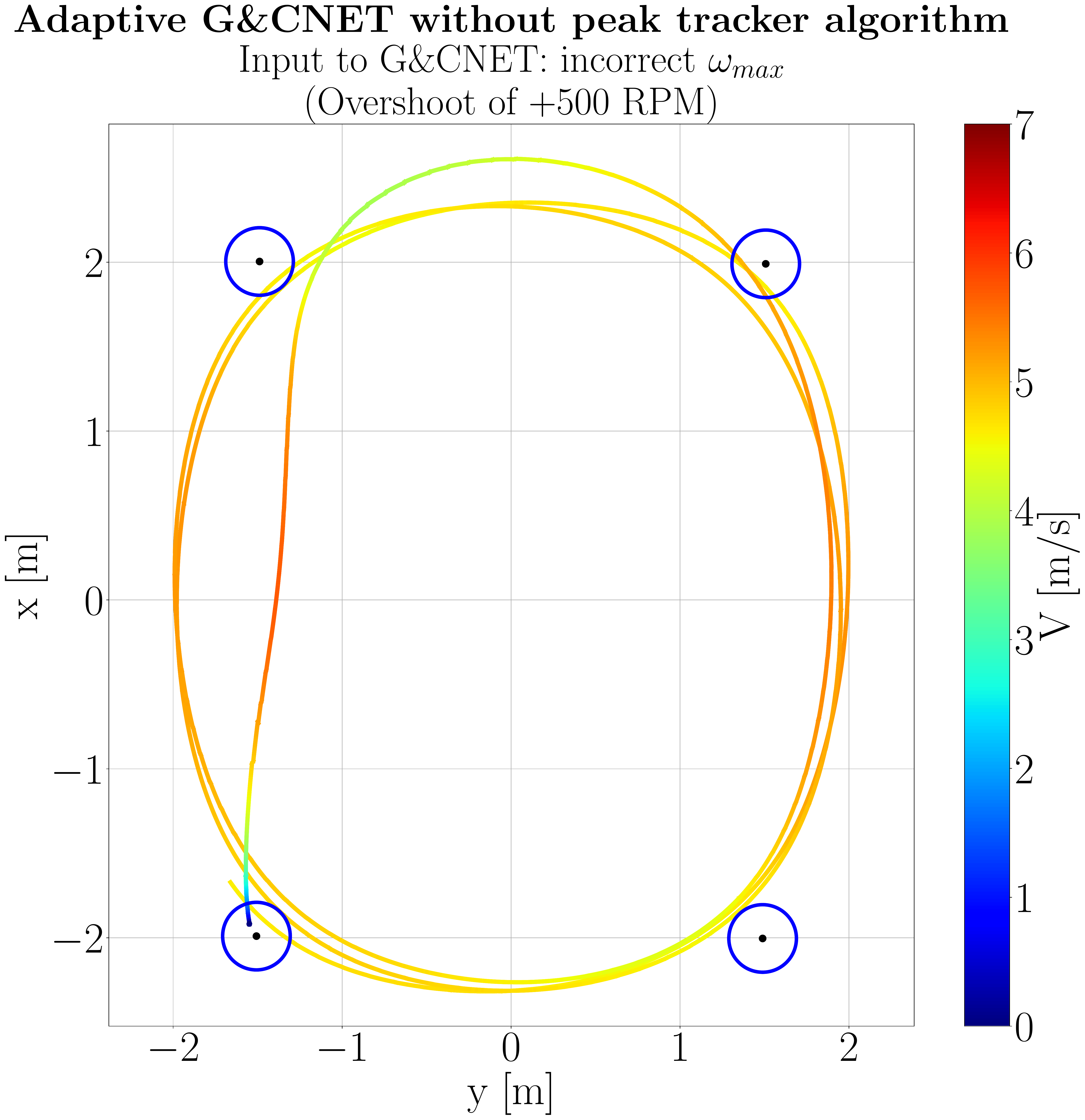}
  \end{minipage}
  \hfill
  \begin{minipage}[b]{0.32\textwidth}
    \includegraphics[width=\textwidth]{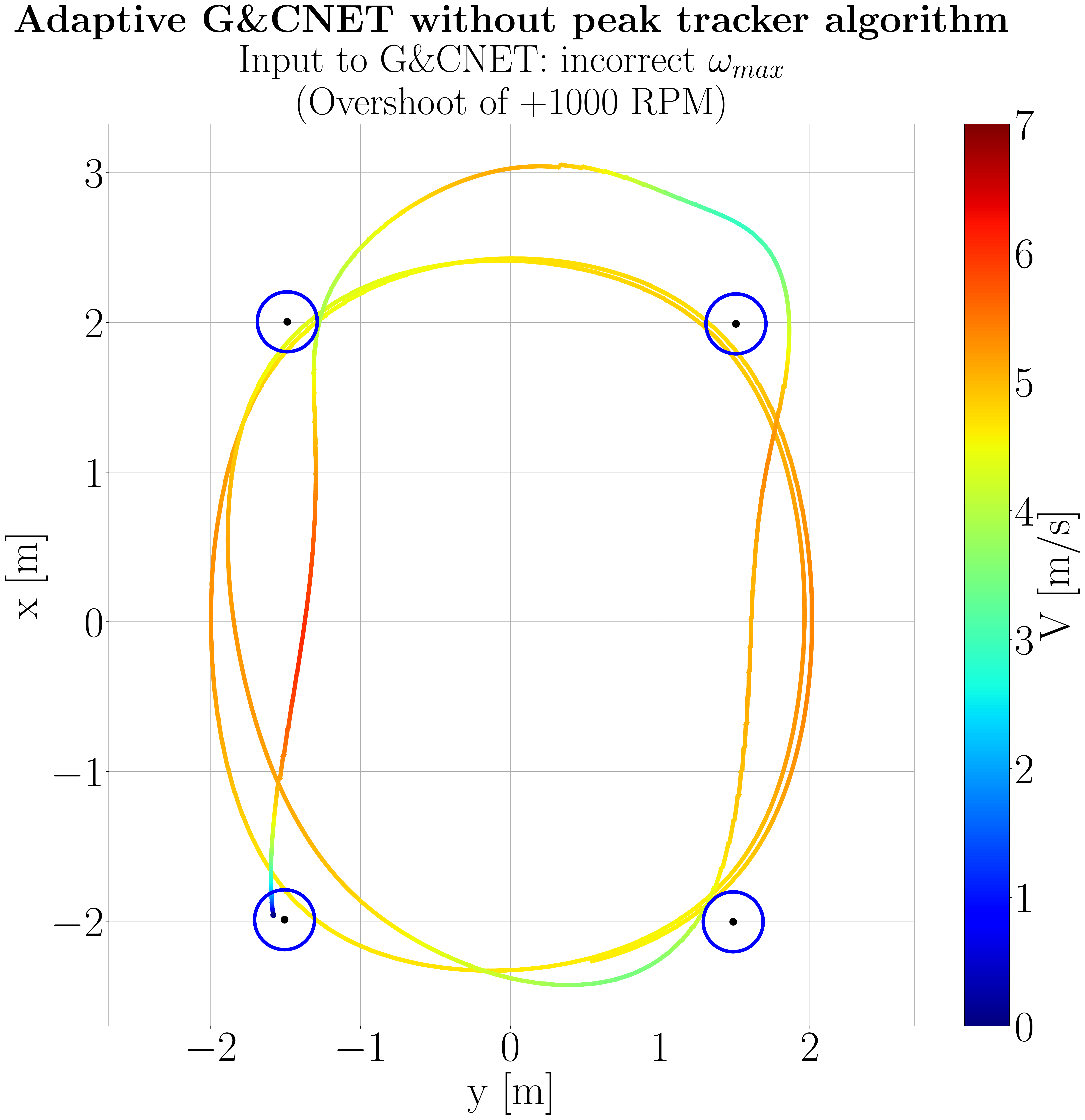}
  \end{minipage}
  \caption{Trajectories (top view) of a real flight with $\epsilon=0.5$. Left: no $\omega_{max}$ overshoot. Center: $\omega_{max}$ overshoot of $+500$ RPM. Right: $\omega_{max}$ overshoot of $+1000$ RPM. During training, the optimal trajectories need to pass within a sphere of radius $20\si{\centi\meter}$ which is indicated by the blue circles, see Sec.\ref{sec:Consecutive WP flight} for details.\vspace{-5mm}}
  
  \label{fig:correct_vs_incorrect_max_rpm}
\end{figure*}

Consider Fig.\ref{fig:correct_vs_incorrect_max_rpm} which shows three real flights with the same G\&CNET ($\epsilon=0.5$). This G\&CNET is trained on 60,000 optimal trajectories and $\omega_{max}$ is sampled uniformly in $[10500,13000]$ RPM. We artificially limit the maximum angular velocity of the propellers to $\omega_{max}=11000\text{ RPM}$. The trajectory on the left shows the resulting flight when the G\&CNET receives the correct $\omega_{max}$ as input, the other two trajectories are flights where the G\&CNET receives an incorrect $\omega_{max}$ as input (overshoot of $+500$ RPM and $+1000$ RPM, respectively). Overshoot refers to the G\&CNET assuming that $\omega_{max}$ is higher than it actually is. The differences between these three cases are mostly visible during the first lap. The aggressive start of the G\&CNET (from hover to pitch down of $\theta=-85\si{\degree}$) and the first turn deviate substantially from the optimal path the more one overshoots $\omega_{max}$.



\begin{figure}[htbp]
  \centering
  \includegraphics[width=\columnwidth]{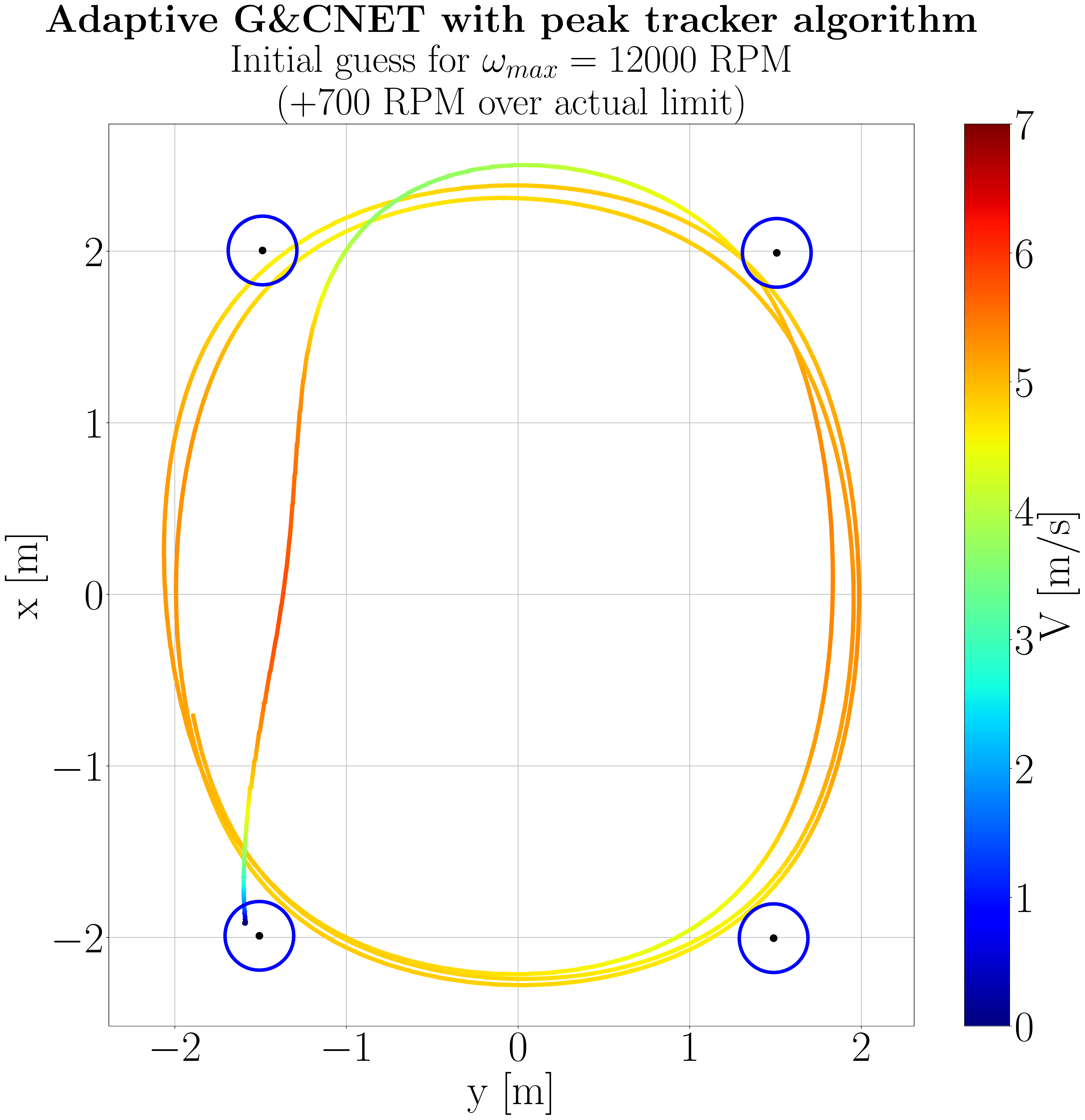}
  \caption{Trajectory (top view) of a real flight with $\epsilon=0.5$ using the peak tracker algorithm and an adaptive G\&CNET. The initial guess for $\omega_{max}$ is off by $+700$ RPM. During training, the optimal trajectories need to pass within a sphere of radius $20\si{\centi\meter}$ which is indicated by the blue circles, see Sec.\ref{sec:Consecutive WP flight} for details.\vspace{-5mm}}
  \label{fig:adaptive_gcnet_with_peak_tracker_algo}
\end{figure}

A way to fly more robustly is to use this same G\&CNET in combination with the peak tracker algorithm. We set the initial guess for maximum RPM to $\omega_{max}=12000$ and we do not artificially limit the maximum angular velocity of
the propellers. In the case of the Bebop 1, the real physical limit of the propellers is $\omega_{max}=11300$, hence we overshoot the limit by $+700$ RPM. It takes the peak tracker $0.2\si{\second}$ after the start of the flight and $0.1\si{\second}$ after the first rotor saturates to identify the correct limit ($\omega_{max}=11300$) and feed it to the G\&CNET. The resulting flight is shown in Fig.\ref{fig:adaptive_gcnet_with_peak_tracker_algo}. Despite initially overshooting the correct limit by $+700$ RPM the quadcopter does not deviate as much from the optimal path as is the case in Fig.\ref{fig:correct_vs_incorrect_max_rpm}.

More research has to be done in this area, it is conceivable that for more aggressive flight (e.g. $\epsilon=0.0$), overestimating $\omega_{max}$ becomes even more critical. Recall Fig.\ref{fig:control_input_diff_max_rpm}, where saturating the rotors $0.1\si{\second}$ too late would result in breaking $1\si{\meter}$ behind the optimal breaking point for a quadcopter travelling at $10\si{\meter\per\second}$. In addition, we learned that the range of values for $\omega_{max}$ one chooses to train a G\&CNET on affects how well the network flies overall. For example, a G\&CNET trained with $\omega_{max}$ uniformly sampled in $[10500,13000]$ RPM will fly less consistent laps than a network with a smaller $\omega_{max}$ range (e.g. $[11000,12000]$ RPM). This might indicate that the current network architecture needs to be revisited to learn the control policies more accurately. One can see in Fig.\ref{fig:correct_vs_incorrect_max_rpm},\ref{fig:adaptive_gcnet_with_peak_tracker_algo} that the G\&CNET struggles to fly through the first waypoint during the first laps.

\section{Consecutive waypoints flight}\label{sec:Consecutive WP flight}

The authors of \cite{time_optimal_planning} have analyzed the gaze of human drone pilots which showed that they looked at multiple gates in advance as opposed to only fixating on the next gate. Previous work \cite{aggressive_online_control, robin_thesis} focused on training and deploying G\&CNET on quadcopters to fly from one point in space to a specific waypoint. By cleverly setting the final conditions of the OCP, it is possible to switch the position of the waypoint right before the quadcopter arrives to make it fly continuously. The obvious next step is to train the G\&CNET on two consecutive waypoints. The global optimal trajectory would optimize for the full track, taking all waypoints into account at one. However, optimizing for a horizon of at least two waypoints will get the drone closer to optimality than a single waypoint approach. We propose one way in which two consecutive waypoints flight can be implemented and show some of the benefits of such a guidance strategy.

\subsection*{Methodology to learn complex tracks}

Consider the task of flying through a set of gates in a time-optimal fashion. The optimal approach to one gate depends heavily on the position and orientation of the next gate. To learn different optimal trajectories based on the relative position and orientation of two consecutive gates we add an intermediate constraint in AMPL. This constraint enforces the optimal solution to pass through a sphere that is centered at $\text{\textbf{WP}}_{1,pos}=[\text{WP}_{1,x}, \text{WP}_{1,y}, \text{WP}_{1,z}]$. Although we are not using the camera of the drone in this work, it is important that the drone's heading $\psi$ always points in a direction such that the next waypoint is in the field of view of the camera. Hence we also add a term to enforce the heading $\text{WP}_{1,psi}$ at the intermediate waypoint:

\begin{equation*} \label{eq:mid_wp_constraint}
\begin{split}
    \text{WP}_{1,threshold} &\ge (x^*_i-\text{WP}_{1,x})^2 + (y^*_i-\text{WP}_{1,y})^2  \\
    & +(z^*_i-\text{WP}_{1,z})^2 +(\psi^*_i - \text{WP}_{1,psi})^2\\
\end{split}
\end{equation*}

\begin{figure}[htbp]
\centering
\tikzset{%
  every neuron/.style={
    circle,
    draw,
    minimum size=0.6cm
  },
  neuron missing/.style={
    draw=none, 
    scale=4,
    text height=0.333cm,
    execute at begin node=\color{black}$\vdots$
  },
}
\resizebox{\columnwidth}{!}{
\begin{tikzpicture}[x=1.4cm, y=1cm, >=stealth]

\foreach \m/\l [count=\y] in {1,2}
  \node [every neuron/.try, neuron \m/.try] (input-\m) at (-1.5,0.75-\y*1.15) {};
\foreach \m [count=\y] in {1,missing,2}
  \node [every neuron/.try, neuron \m/.try ] (hidden-\m) at (-1.0,1.5-\y*1.25) {};
\foreach \m [count=\y] in {1,missing,2}
  \node [every neuron/.try, neuron \m/.try ] (hidden2-\m) at (0.2,1.5-\y*1.25) {};
 \foreach \m [count=\y] in {1,missing,2}
  \node [every neuron/.try, neuron \m/.try ] (hidden3-\m) at (1.4,1.5-\y*1.25) {};
\foreach \m [count=\y] in {1,2,3,4}
  \node [every neuron/.try, neuron \m/.try ] (output-\m) at (2.6,1.5-\y*1.0) {};

\foreach \l [count=\i] in {\mb{x}, \mb{WP}_{rel}}
  \draw [<-] (input-\i) -- ++(-0.95,0)
    node [above, midway] {$\l$};
\foreach \l [count=\i] in {1,2,3,4}
  \draw [->] (output-\i) -- ++(0.7,0)
    node [above, midway] {$u_\l$};

\foreach \i in {1,2}
  \foreach \j in {1,...,2}
    \draw [->] (input-\i) -- (hidden-\j);
\foreach \i in {1,...,2}
  \foreach \j in {1,...,2}
    \draw [->] (hidden-\i) -- (hidden2-\j);
\foreach \i in {1,...,2}
  \foreach \j in {1,...,2}
    \draw [->] (hidden2-\i) -- (hidden3-\j);
\foreach \i in {1,...,2}
  \foreach \j in {1,2,3,4}
    \draw [->] (hidden3-\i) -- (output-\j);

\foreach \l [count=\x from 1] in {(ReLU)\\120\\neurons, (ReLU)\\120\\neurons,(ReLU)\\120\\neurons, (Sigmoid) \\ Output\\layer}
  \node [align=center, above] at (-2.2+\x*1.2,1) {\l};
\end{tikzpicture}
}
\caption{Feed forward network architecture for G\&CNETs using multiple waypoints. We add the input $\mb{WP}_{rel}$ which informs the network on the relative position of the two upcoming waypoints.\vspace{-0mm}}
\label{fig:ffnn_consecutive_wps}
\end{figure}
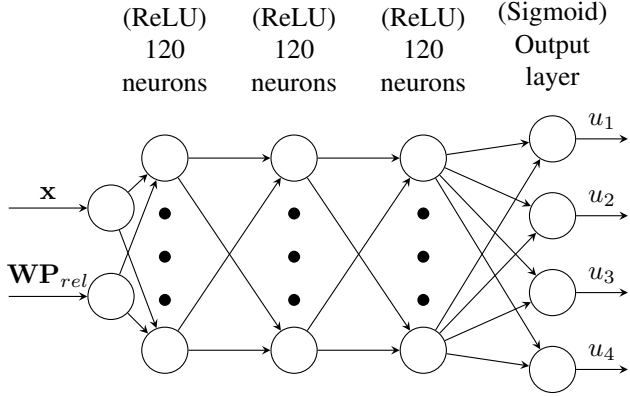

We choose this constraint because it allows to relax the position and heading error at the intermediate waypoint. It is difficult to know in advance what the optimal path and heading at any given stage of a complex track should be. Implementing the constraint this way gives the solver some freedom to pass through the intermediate waypoint in a more optimal way based on the initial conditions and the relative position of the two waypoints. The bottom figure in Fig.\ref{fig:opt_traj_training} (App.\ref{app:opt_traj_training}) shows optimal trajectories with the intermediate waypoint constraint.
Since the trajectories are now on average longer than for single waypoint flight we sample them in $N=319$ points. We inform the G\&CNET on the relative position of the two upcoming waypoints $\mb{WP}_{rel}$ by adding it as an input to the network architecture, see Fig.\ref{fig:ffnn_consecutive_wps}. We do not add $\omega_{max}$ as additional input to the network architecture here as the size of the network is reaching its limit in terms of the amount of information it can carry. In order to always reach the desired saturation level, we fly conservatively at an $\omega_{max}$ that the quadcopter can reach.

This methodology lends itself well to learning to fly more complex tracks as different types of optimal turns can be generated. In Appendix \ref{app:complex_track} we provide a trajectory of flight in simulation on a figure-eight track. The training dataset can either be tailored to a track if it is known in advance or as we will see at the end of this section, it is also possible to cover a range of relative positions between the two waypoints such that the G\&CNET can fly different variations of tracks.


\subsection*{Comparison with energy-optimal single waypoint flight}
\begin{figure*}[hbtp]
  \centering
  \begin{minipage}[b]{0.333\textwidth}
    \includegraphics[width=\textwidth]{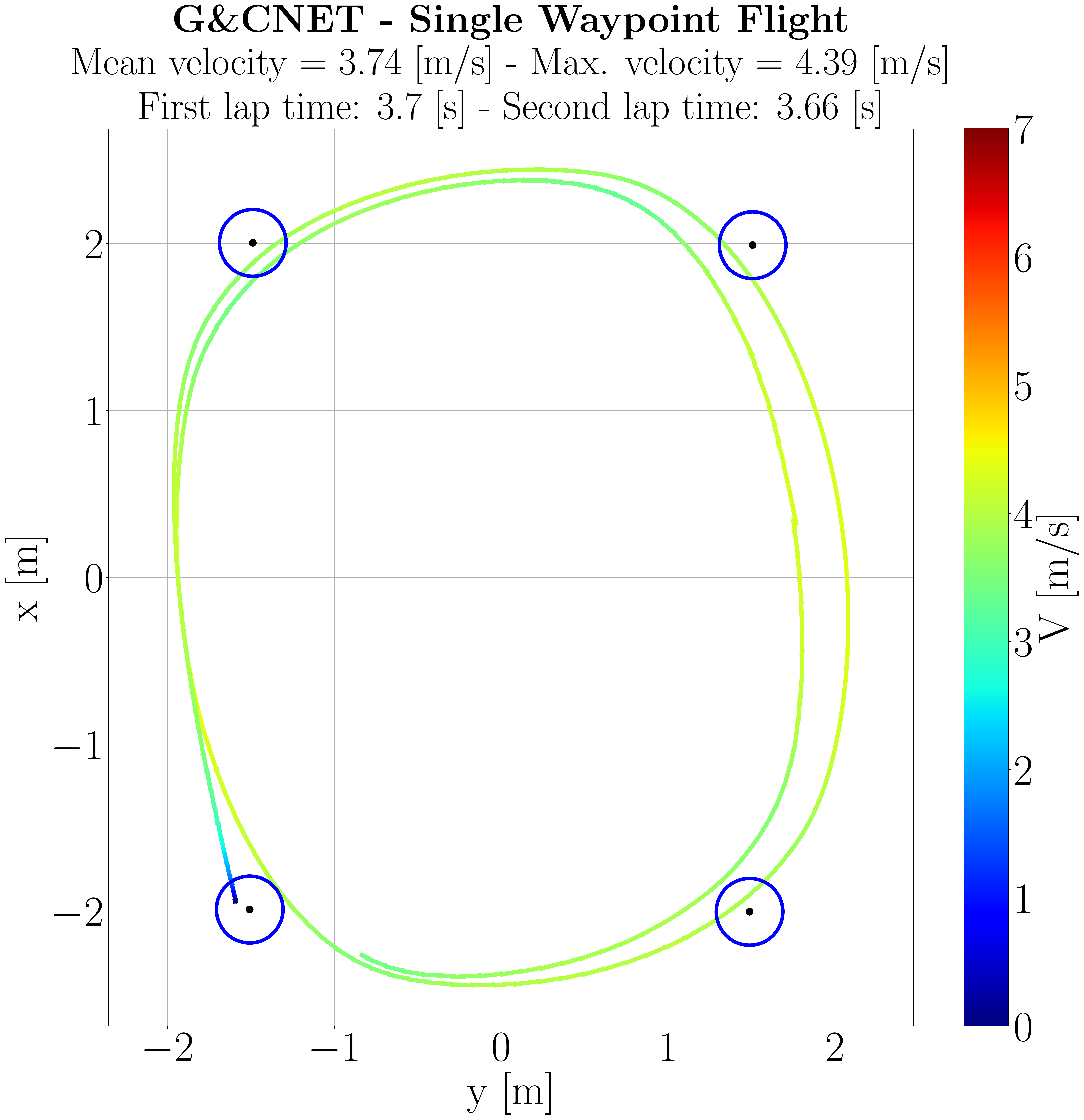}
  \end{minipage}
  \hfill
  \begin{minipage}[b]{0.333\textwidth}
    \includegraphics[width=\textwidth]{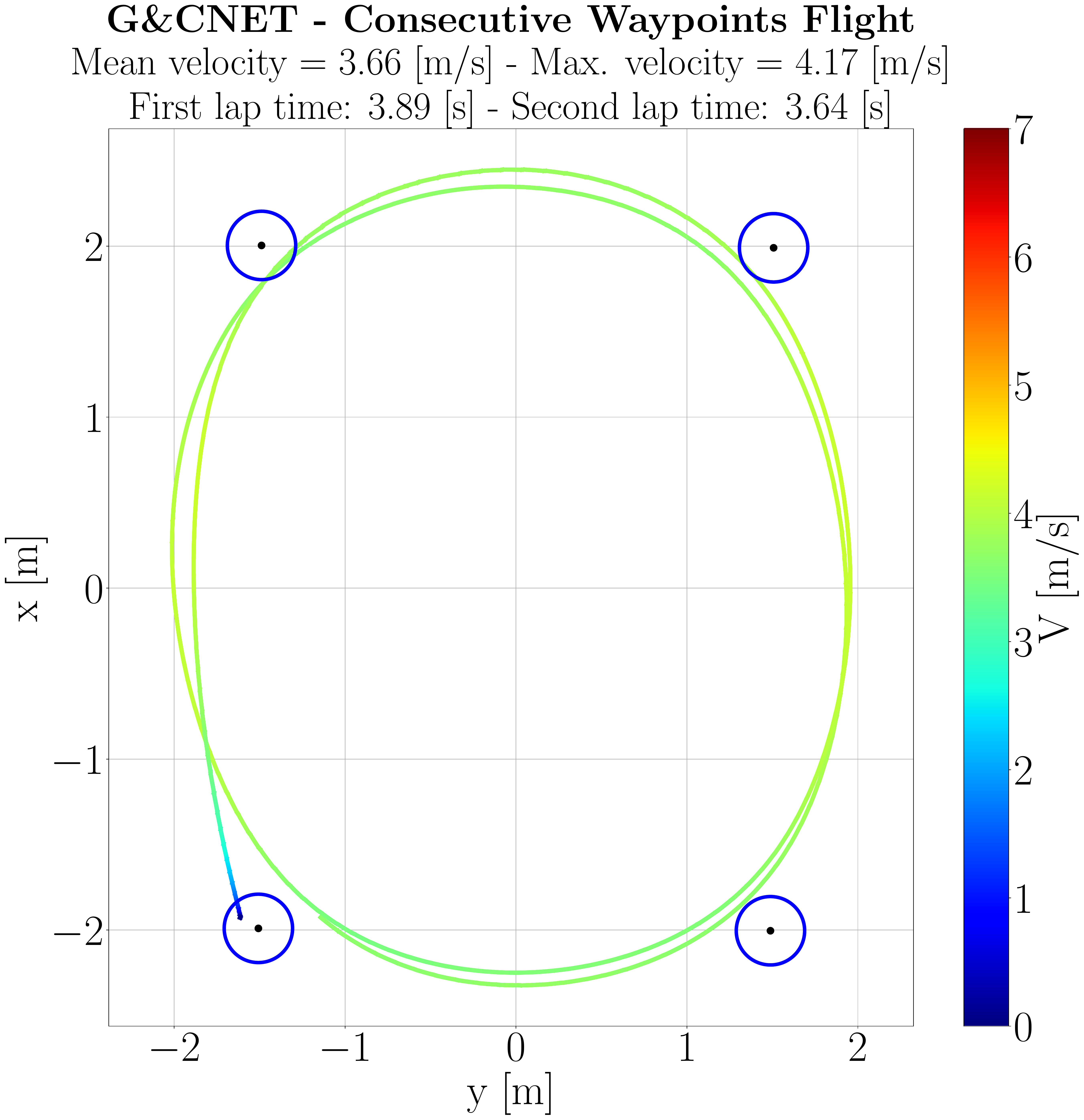}
  \end{minipage}
  \hfill
  \begin{minipage}[b]{0.3\textwidth}
    \includegraphics[width=\textwidth]{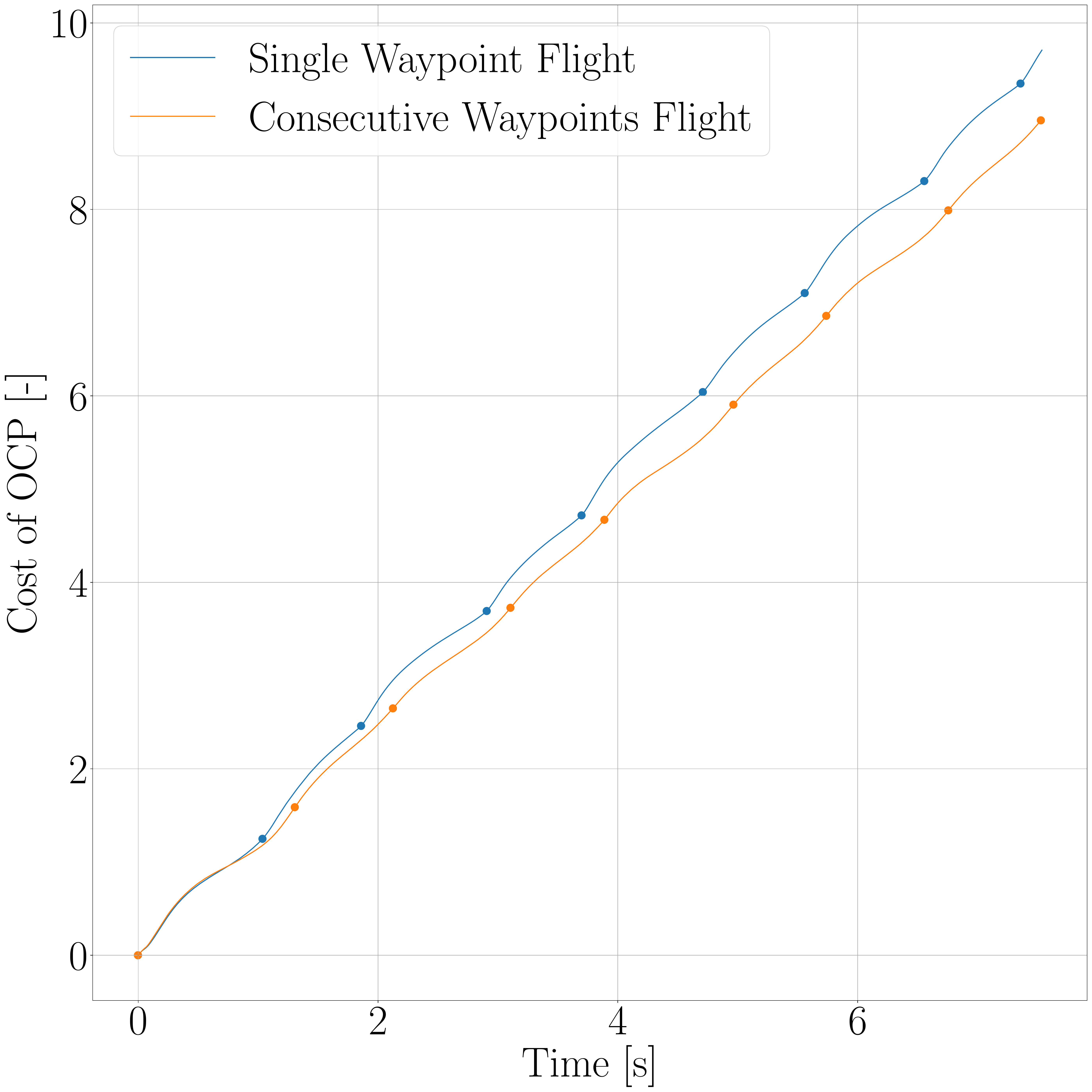}
  \end{minipage}
  \caption{Trajectories (top view) of a real flight with energy-optimal G\&CNETs. Left: single waypoint flight. Center: consecutive waypoints flight. Right: cost function over time for both flights. The dots indicate the switching times to the next waypoint(s). The blue circles indicate the position constraint at the intermediate waypoint.\vspace{-2mm}}
  \label{fig:single_and_consecutive_wps_flight_energy_opt_and_cost}
\end{figure*}

Another advantage of learning to fly while taking more than one waypoint into account is that the G\&CNET flies more optimally through the entire track. We show this by performing two real flights on the Bebop with energy-optimal G\&CNETs and subsequently computing the cost (Eq.\ref{eq:cost_function} with $\epsilon=1.0$) over time. We choose the energy-optimal control problem here because it is the easiest optimal control policy for G\&CNETs to learn. The same analysis could also have been done for a different $\epsilon$. Fig.\ref{fig:single_and_consecutive_wps_flight_energy_opt_and_cost} shows the resulting trajectories. The left figure shows the G\&CNET which is only trained on a single waypoint. We switch to the next waypoint when the Euclidean distance in three dimensions between the quadcopter and the waypoint is below $1.2\si{\meter}$ \cite{robin_thesis}. We experimented with varying switching distances. In general, the larger the switching distance, the more the quadcopter will cut the corner. For smaller switching distances the position error from the waypoint becomes smaller, however, the quadcopter slows down as the G\&CNET tries to perfectly meet the final conditions of the OCP. Note that we use a free final magnitude in velocity, only the direction of the velocity vector is constrained to be aligned with the desired final heading angle ($\psi_f = 45^\circ$). The other G\&CNET (center figure) is trained on two consecutive waypoints with $\text{WP}_{1,threshold}=0.2$ and $\text{WP}_{1,psi} =45^\circ$. This means that optimal trajectories have to pass through a sphere of radius $20\si{\centi\meter}$ with a heading of roughly $\psi=45^\circ$. 
The G\&CNET is trained on a dataset that contains two "types" of turns: trajectories where the two waypoints are $3\si{\meter}$ apart and trajectories where they are $4\si{\meter}$ apart. Hence only one extra input to the G\&CNET is required ($\mb{WP}_{rel}$ in Fig.\ref{fig:ffnn_consecutive_wps}) to inform the network on which of the two possible turns is upcoming. 
A visualization of optimal trajectories used to train G\&CNETs in the case of single and consecutive waypoints is provided in App.\ref{app:opt_traj_training}. Compared to the single waypoint flight, we do not have to trade off 
position errors from the waypoint and speed anymore. We switch waypoints every time the G\&CNET passes one, hence $3\si{\meter}$ before the final waypoint during the first turn and $4\si{\meter}$ before the final waypoint the next turn and so forth. We compute the cost $\int_{0}^{T} ||\mb{u}(t)||^2 dt$ over these flights and plot it in the right-most figure (Fig.\ref{fig:single_and_consecutive_wps_flight_energy_opt_and_cost}). The network that is trained on two consecutive waypoints spends less energy, hence minimizing the cost function better over time. We also note that the control inputs are smoother and saturate less in the case of consecutive waypoints flight, leaving more control authority to recover from errors. This is likely because the G\&CNET trained on a single waypoint unnecessarily saturates the rotors because it decelerates and accelerates more.


\subsection*{Comparison with minimum snap benchmark}

\begin{figure*}[!tbp] 
  \centering
  \begin{minipage}[b]{0.49\textwidth}
    \includegraphics[width=\textwidth]{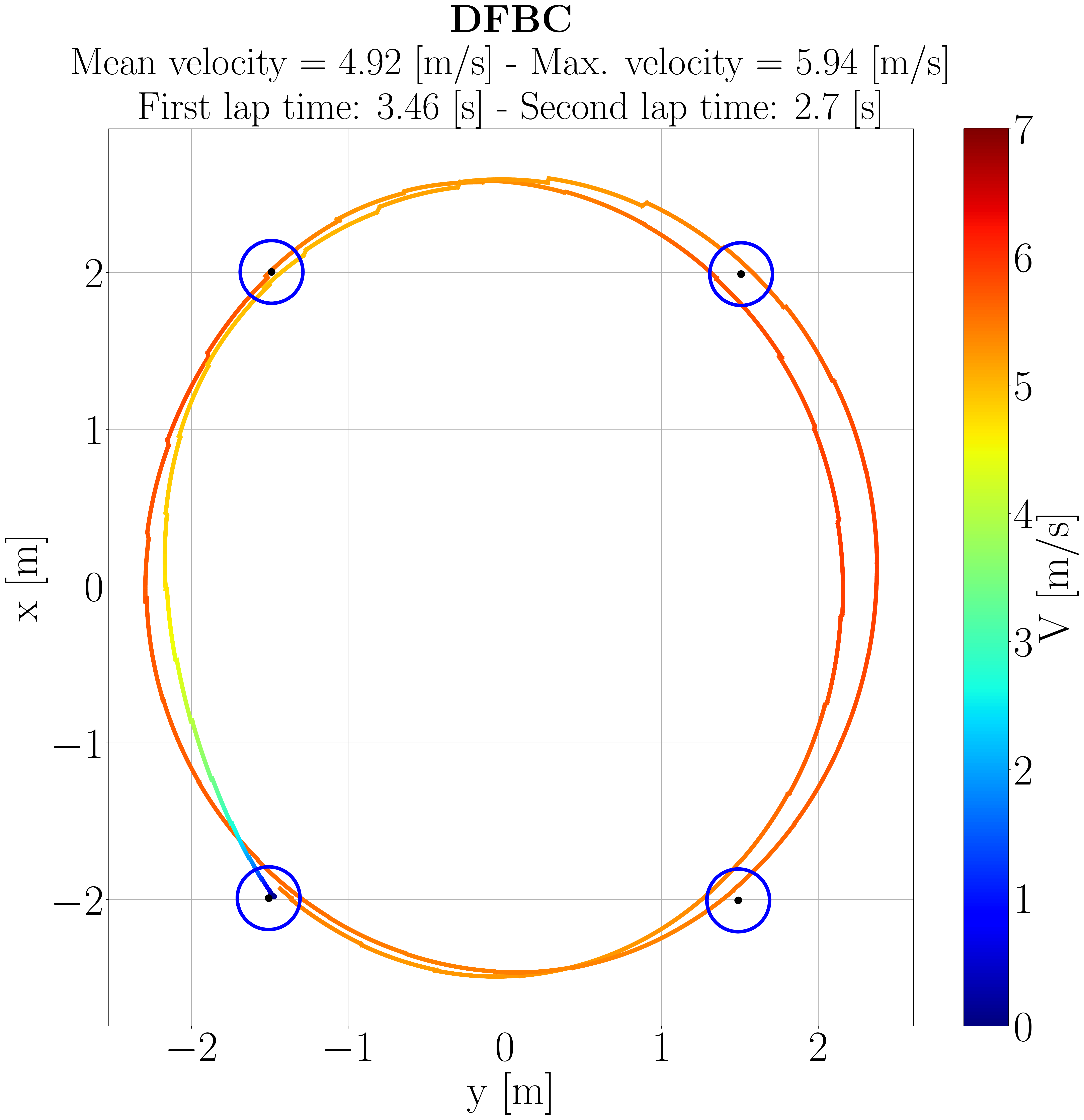}
  \end{minipage}
  \hfill
  \begin{minipage}[b]{0.48\textwidth}
    \includegraphics[width=\textwidth]{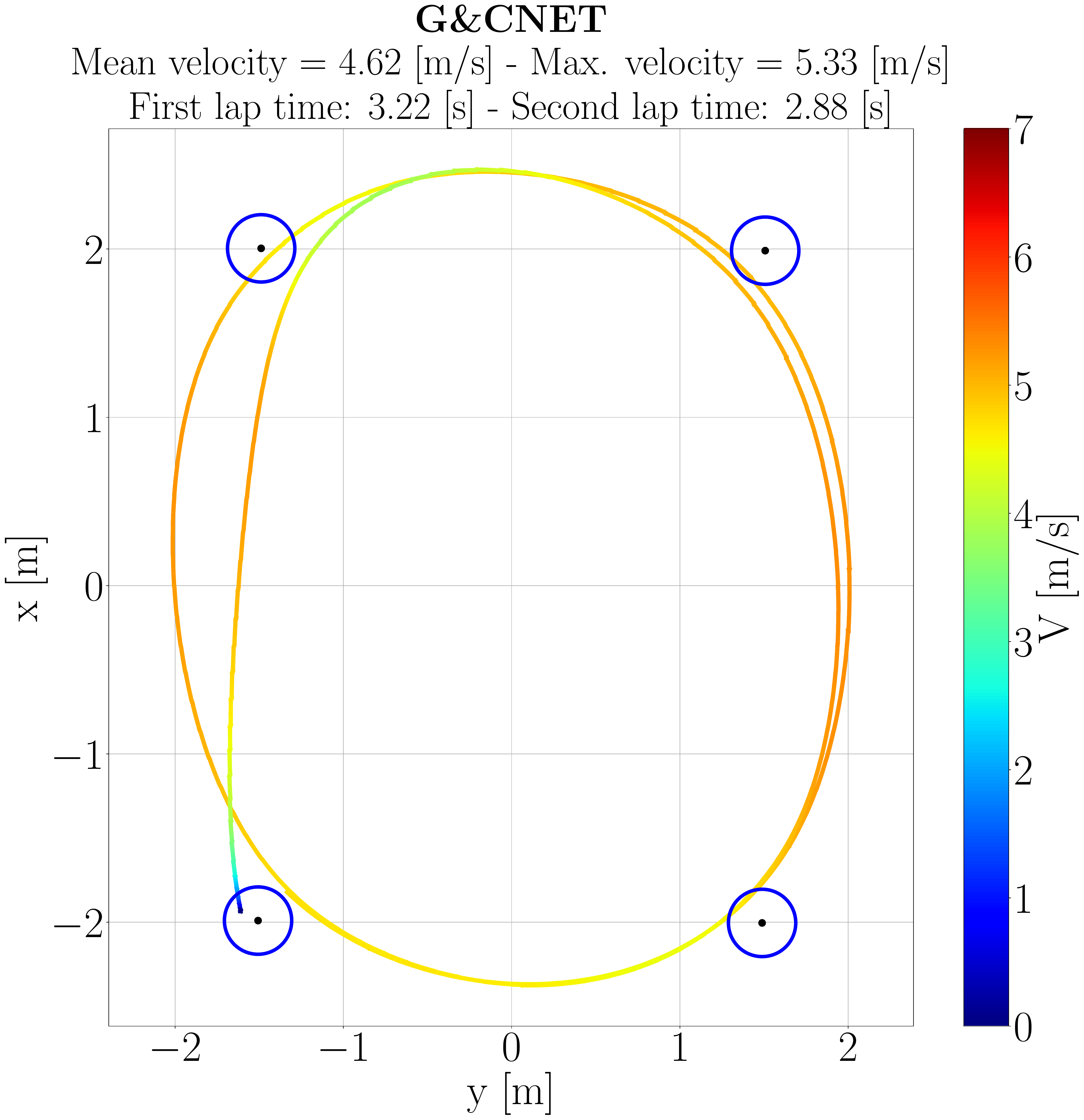}
  \end{minipage}
  \caption{Left: trajectory (top view) of a real flight with a DFBC (min snap). Right: trajectory (top view) of a real flight  $\epsilon=0.5$. The G\&CNET is trained on 2 consecutive waypoints. The blue circles indicate the position constraint at the intermediate waypoint for the G\&CNET.\vspace{-5mm}}
  \label{fig:fastest_gcnet}
\end{figure*}

We also consider flying the $4\times3\si{\meter}$ track as fast as possible with the G\&CNET. We choose the well-known differential-flatness-based-controller (DFBC) \cite{MinSnap} as a benchmark to compare our G\&CNET. This state-of-the-art controller uses polynomials to generate smooth trajectories by minimizing snap, the fourth derivative of position. The reference trajectory is then tracked by an outer-loop Incremental Nonlinear Dynamic Inversion (INDI) controller. 
Just as for the energy-optimal case, the G\&CNET is trained on trajectories where the two waypoints are $3\si{\meter}$ apart and trajectories where they are $4\si{\meter}$ apart, only now $\epsilon=0.5$.
The resulting flights are shown in Fig.\ref{fig:fastest_gcnet}. Since the G\&CNET always tries to fly through the apex of the blue circles in Fig.\ref{fig:fastest_gcnet}, we moved the waypoints in the polynomial generation for the DFBC inwards such that they coincide with these apexes to make the comparison fairer. Both controllers have their advantages, it is for instance possible to make more aggressive maneuvers with the G\&CNET. This is especially noticeable in the first lap which the G\&CNET performs in $3.22\si{\second}$ (DFBC takes $3.46\si{\second}$). The DFBC however can sustain higher velocities once the transient behaviour at the start is over. The DFBC performs the second lap in $2.7\si{\second}$ compared to $2.88\si{\second}$ for the G\&CNET. The commanded and observed angular velocities during these flights are provided in App.\ref{app:control_inputs_dfbc_gcnet}. The control inputs of the G\&CNET are considerably smoother than the ones for the DFBC. One should note here that both controllers still have room for improvement. Weighted least squares could be implemented in the control allocation for the INDI used with the DFBC. In addition, the G\&CNET has an unfair advantage in this comparison as it is trained for $\omega_{max}=11300$ RPM which is close to the true limit of the Bebop. The DFBC however assumes that $\omega_{max}=12000$ RPM. Finally, it is possible to fly faster with the G\&CNET (lower $\epsilon$) at the expense of larger position errors and more unstable flight.



\subsection*{Flexibility of G\&CNETs}\label{subsec:rand_wps}

Finally, we highlight a major of advantage of G\&CNET over the DFBC: the ability to recompute trajectories and the corresponding optimal controls online. G\&CNETs are very flexible in so far as new optimal controls $\mb{u}^*$ are immediately computed even when deviating from the globally optimal path (so long as the state of the quadrotor has been represented closely enough in the training data set). This also means that the G\&CNET can handle different waypoint positions within the training data. The DFBC on the other hand can only rely on the trajectory which has been generated offline. The DFBC will always try to stay as close as possible to this trajectory. This is a problem as deviations from this trajectory are bound to happen due to the reality gap. Once deviated this trajectory is no longer optimal and a new one should be computed. Moving a waypoint also requires computing a new optimal trajectory offline. Finally, due to its relatively small network size, the G\&CNET can be inferred onboard the Bebop at a frequency of $450$ Hz with the current network architecture. Both of these characteristics allow G\&CNETs to cancel out approximations errors as opposed to accumulating these over time. 

\begin{figure*}[!tbp]
  \centering
  \begin{minipage}[b]{0.49\textwidth}
    \includegraphics[width=\textwidth]{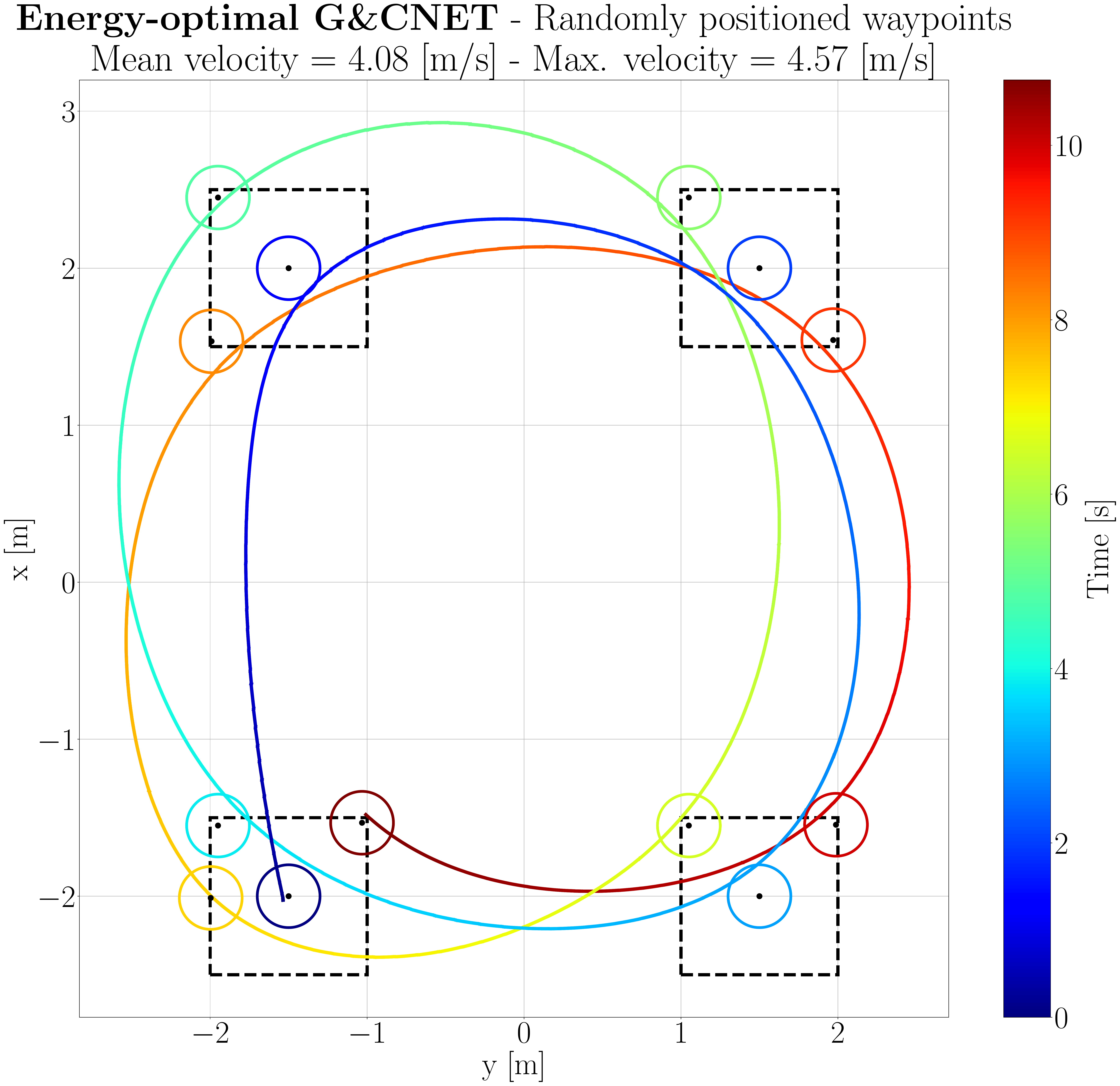}
  \end{minipage}
  \hfill
  \begin{minipage}[b]{0.48\textwidth}
    \includegraphics[width=\textwidth]{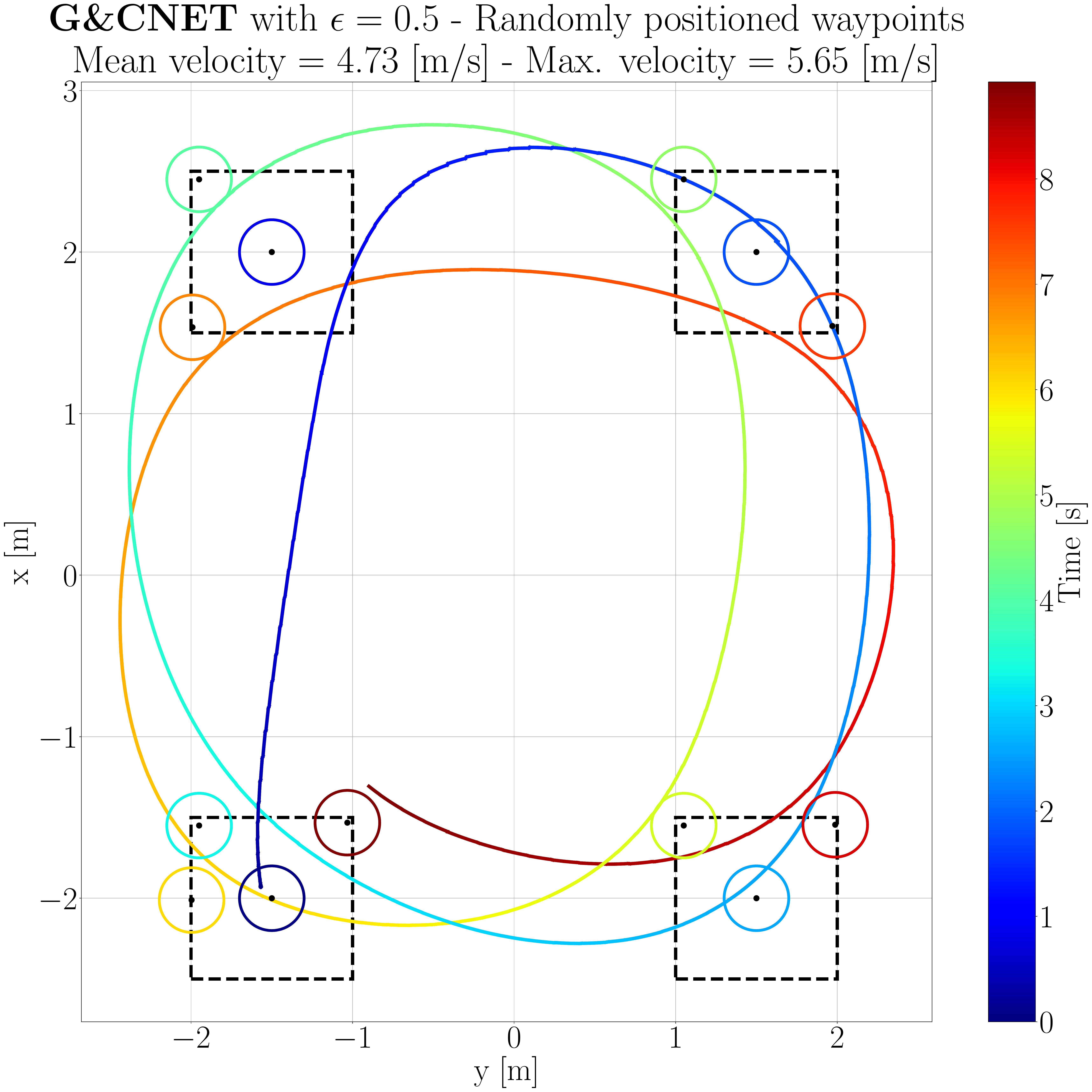}
  \end{minipage}
  \caption{Trajectories (top view) of a real flights with $\epsilon=1.0$ (left) and $\epsilon=0.5$ (right). The G\&CNETs are trained on 2 consecutive waypoints with varying relative positions. We randomly position the waypoints within the dashed rectangles. The circles indicate the position constraint for each waypoint.\vspace{-2mm}}
  \label{fig:randomly_pos_wps}
\end{figure*}

We consider the task of flying the $4\times3\si{\meter}$ track, however, the four waypoints are now randomly moved (before the G\&CNET takes them into account) within a square of $1\si{\meter\squared}$ in the XY plane (see dashed squares in Fig.\ref{fig:randomly_pos_wps}). We train a network on a range of relative waypoint positions by uniformly sampling $\text{WP}_{1,x}, \text{WP}_{1,y}$ in a square of $1\si{\meter\squared}$ centered at $3.5\si{\meter}$ from the final waypoint. The altitude is kept constant in this experiment for simplicity. Since the relative waypoint position can now vary in two dimensions, two extra inputs are required ($\mb{WP}_{rel}$ in Fig.\ref{fig:ffnn_consecutive_wps}) to inform the network. We train two G\&CNETs (one energy-optimal and one with $\epsilon=0.5$) and fly these on the Bebop, the resulting trajectories are shown in Fig.\ref{fig:randomly_pos_wps}. In both cases, the networks manage to adapt their trajectory. The position errors from the waypoint become notably larger for the faster G\&CNET as the policy is more difficult to learn (Sec.\ref{sec:Time-optimal quadcopter flight}) and the reality gap (hardware delays, state estimation errors and modelling errors) are harder to cope with when one flies more time-optimally.

\begin{figure}[htbp] 
  \centering
  \includegraphics[width=\columnwidth]{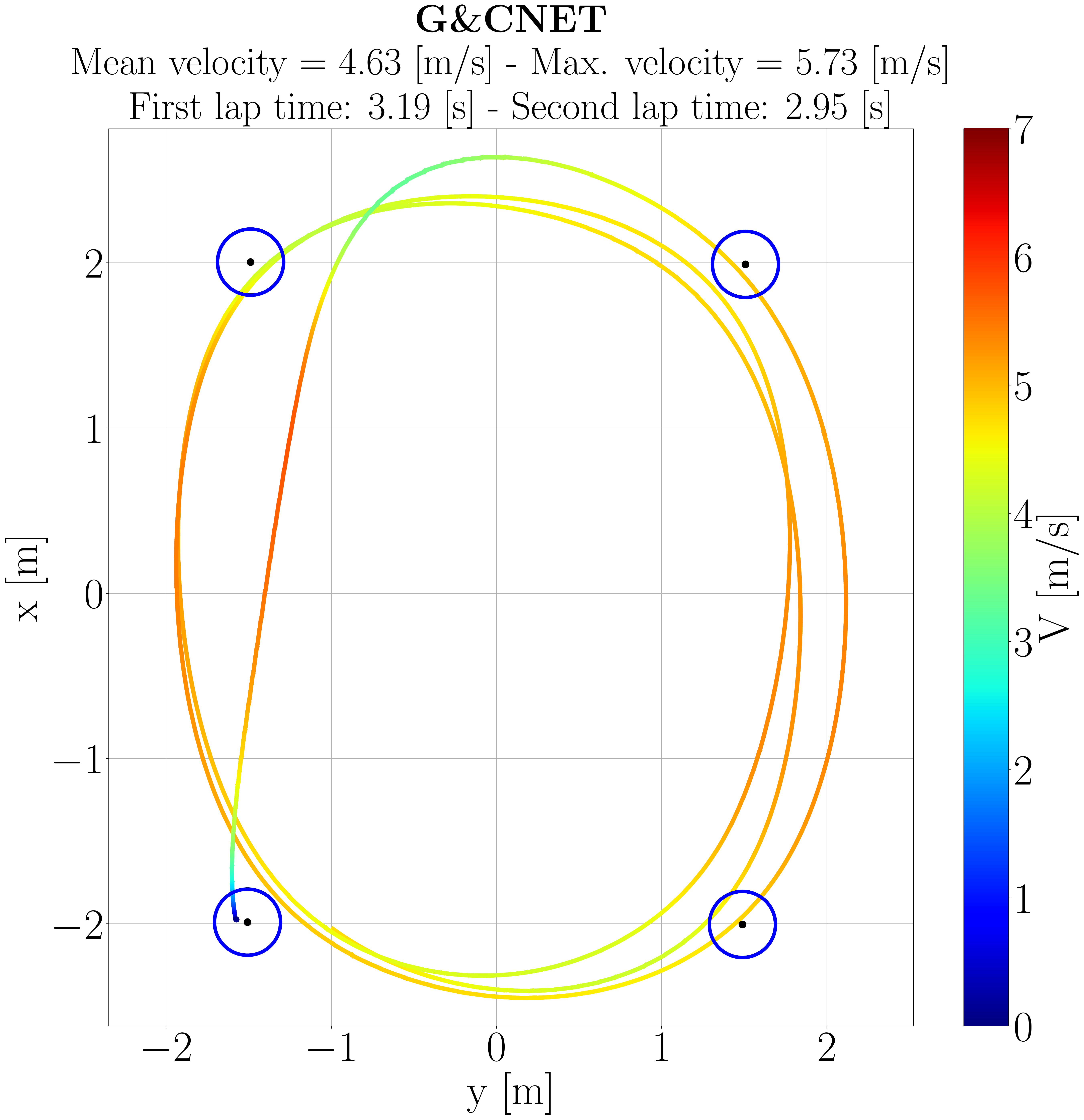}
  \caption{Trajectory (top view) of a real flight $\epsilon=0.5$. The G\&CNET is trained on 2 consecutive waypoints with varying relative positions. The blue circles indicate the position constraint at the intermediate waypoint.\vspace{-5mm}}
  \label{fig:gcnet_varying_wps}
\end{figure}

A similar observation as in Sec.\ref{sec:Accounting for the varying max RPM limit of quadcopters} is made here regarding lower control accuracy when training the G\&CNET on a larger range of data. We flew the $4\times3\si{\meter}$ track with fixed waypoints using the G\&CNET that has learned to fly on a range of different waypoint positions, the G\&CNET flies less consistent laps (Fig.\ref{fig:gcnet_varying_wps}) compared to the G\&CNET that is specifically trained on the $4\times3\si{\meter}$ track (Fig.\ref{fig:fastest_gcnet}). Both networks roughly have the same network architecture (only one extra input neuron for the case where both $\text{WP}_{1,x}$ and $\text{WP}_{1,y}$ are varied). It is possible that the current network size needs to be increased for both G\&CNETs to fly the $4\times3\si{\meter}$ track with the same accuracy.

\section{Conclusion}
Guidance \& Control Networks have been studied in the context of fast quadcopter flight. We showed that the control policies for the time-optimal control problem are considerably more difficult to learn than for the energy-optimal control problem. For close to time-optimal flight with the Bebop 1, average control errors of $\pm3.35\%$ are already too high to maintain stable flight in simulation. We demonstrated that the maximum angular speed of propellers $\omega_{max}$ affects the switching times in the time-optimal control profile. We then went on to show that the more one over- or underestimates $\omega_{max}$, the larger the mean the position error from the optimal trajectory becomes, which in turn affects the robustness of the flight. We propose a peak tracker algorithm to identify $\omega_{max}$ onboard in combination with a G\&CNET that can adapt its control policy based on the identified value for $\omega_{max}$. Our algorithm takes $0.1\si{\second}$ after one of the four rotors saturates to identify the new limit, allowing it to stay close to the optimal trajectory in a 
real flight even when initially overestimating $\omega_{max}$ by $+700$ RPM. Finally, we extend previous work on G\&CNETs by learning to fly while taking two upcoming waypoints into account. The new pipeline allows to generate training datasets that contain specific maneuvers for the G\&CNET to learn, allowing it for instance to fly a figure-eight track in simulation. Compared to single-waypoint flight we optimize the energy-optimal cost function better over a $4\times3\si{\meter}$ track since the OCP formulation for multiple waypoints is more representative of the entire control task. We considered flying the $4\times3\si{\meter}$ track as fast as possible and benchmarking our G\&CNET against the state-of-the-art differential-flatness-based-controller (DFBC). We show that G\&CNETs can fly the track in similar lap times as the DFBC and  adapt to varying waypoint positions. This highlights one of the main advantages of G\&CNET compared to other optimality-based approaches, such as the DFBC: its flexibility to quickly recompute optimal control inputs.

Future work can be done on identifying the maximum angular velocity of each individual rotor, thereby not restricting propellers that are experiencing less aerodynamic load than the most limiting propeller. A more rigorous constraint could be implemented to make sure the waypoints are always in the field of view of the camera. The current dynamic model of the Bebop does not include the effects of downwash and errors in the thrust and drag model are common for flights where $\epsilon<0.5$, hence one could consider using domain randomization in combination with onboard measurements to adapt to these model inaccuracies during flight. Finally, the methodology used in this work for two consecutive waypoints could be used to train a G\&CNET on a much larger range of possible waypoint combinations, thereby allowing the network to fly a lot of different tracks.

    



\section*{Acknowledgment}
The authors are grateful to Emmanuel Blazquez and Alexander Hadjiivanov for their valuable inputs and discussions throughout this project and to Erik van der Horst for his technical support. This research was co-funded under the Discovery programme of, and funded by, the European Space Agency.

\bibliographystyle{IEEEtran}
\bibliography{references}

\begin{thebibliography}{10}
\providecommand{\url}[1]{#1}
\csname url@samestyle\endcsname
\providecommand{\newblock}{\relax}
\providecommand{\bibinfo}[2]{#2}
\providecommand{\BIBentrySTDinterwordspacing}{\spaceskip=0pt\relax}
\providecommand{\BIBentryALTinterwordstretchfactor}{4}
\providecommand{\BIBentryALTinterwordspacing}{\spaceskip=\fontdimen2\font plus
\BIBentryALTinterwordstretchfactor\fontdimen3\font minus
  \fontdimen4\font\relax}
\providecommand{\BIBforeignlanguage}[2]{{%
\expandafter\ifx\csname l@#1\endcsname\relax
\typeout{** WARNING: IEEEtran.bst: No hyphenation pattern has been}%
\typeout{** loaded for the language `#1'. Using the pattern for}%
\typeout{** the default language instead.}%
\else
\language=\csname l@#1\endcsname
\fi
#2}}
\providecommand{\BIBdecl}{\relax}
\BIBdecl

\bibitem{range_optimal_speed_quads}
\BIBentryALTinterwordspacing
L.~Bauersfeld and D.~Scaramuzza, ``Range, endurance, and optimal speed
  estimates for multicopters,'' 2021. [Online]. Available:
  \url{https://arxiv.org/abs/2109.04741}
\BIBentrySTDinterwordspacing

\bibitem{alphapilot_win_mavlab}
C.~{De Wagter}, F.~Paredes-Vall{\'e}s, N.~Sheth, and G.~{de Croon},
  ``\BIBforeignlanguage{English}{Learning fast in autonomous drone racing},''
  \emph{\BIBforeignlanguage{English}{Nature Machine Intelligence}}, vol.~3,
  no.~10, p. 923, 2021, copyright: Copyright 2021 Elsevier B.V., All rights
  reserved.

\bibitem{estimation_control_planning}
G.~Loianno, C.~Brunner, G.~McGrath, and V.~Kumar, ``Estimation, control, and
  planning for aggressive flight with a small quadrotor with a single camera
  and imu,'' \emph{IEEE Robotics and Automation Letters}, vol.~2, no.~2, pp.
  404--411, 2017.

\bibitem{mpccontouring}
\BIBentryALTinterwordspacing
A.~Romero, S.~Sun, P.~Foehn, and D.~Scaramuzza, ``Model predictive contouring
  control for time-optimal quadrotor flight,'' 2021. [Online]. Available:
  \url{https://arxiv.org/abs/2108.13205}
\BIBentrySTDinterwordspacing

\bibitem{Mohta_2017}
K.~Mohta, M.~Watterson, Y.~Mulgaonkar, S.~Liu, C.~Qu, A.~Makineni, K.~Saulnier,
  K.~Sun, A.~Zhu, J.~Delmerico, K.~Karydis, N.~Atanasov, G.~Loianno,
  D.~Scaramuzza, K.~Daniilidis, C.~Taylor, and V.~Kumar, ``Fast, autonomous
  flight in gps-denied and cluttered environments,'' \emph{Journal of Field
  Robotics}, vol.~35, 12 2017.

\bibitem{autonomous_drone_race}
\BIBentryALTinterwordspacing
S.~Li, M.~M. Ozo, C.~{De Wagter}, and G.~C. {de Croon}, ``Autonomous drone
  race: A computationally efficient vision-based navigation and control
  strategy,'' \emph{Robotics and Autonomous Systems}, vol. 133, p. 103621,
  2020. [Online]. Available:
  \url{https://www.sciencedirect.com/science/article/pii/S0921889020304619}
\BIBentrySTDinterwordspacing

\bibitem{mellinger_aggressive_man}
\BIBentryALTinterwordspacing
D.~Mellinger, N.~Michael, and V.~Kumar, ``Trajectory generation and control for
  precise aggressive maneuvers with quadrotors,'' \emph{The International
  Journal of Robotics Research}, vol.~31, no.~5, pp. 664--674, 2012. [Online].
  Available: \url{https://doi.org/10.1177/0278364911434236}
\BIBentrySTDinterwordspacing

\bibitem{drone_acrobatics}
E.~Kaufmann, A.~Loquercio, R.~Ranftl, M.~M\"{u}ller, V.~Koltun, and
  D.~Scaramuzza, ``Deep drone acrobatics,'' \emph{RSS: Robotics, Science, and
  Systems}, 2020.

\bibitem{time_optimal_planning}
\BIBentryALTinterwordspacing
P.~Foehn, A.~Romero, and D.~Scaramuzza, ``Time-optimal planning for quadrotor
  waypoint flight,'' \emph{Science Robotics}, vol.~6, no.~56, p. eabh1221,
  2021. [Online]. Available:
  \url{https://www.science.org/doi/abs/10.1126/scirobotics.abh1221}
\BIBentrySTDinterwordspacing

\bibitem{neuralmpc}
\BIBentryALTinterwordspacing
T.~Salzmann, E.~Kaufmann, M.~Pavone, D.~Scaramuzza, and M.~Ryll, ``Neural-mpc:
  Deep learning model predictive control for quadrotors and agile robotic
  platforms,'' 2022. [Online]. Available:
  \url{https://arxiv.org/abs/2203.07747}
\BIBentrySTDinterwordspacing

\bibitem{MinSnap}
D.~Mellinger and V.~Kumar, ``Minimum snap trajectory generation and control for
  quadrotors,'' in \emph{2011 IEEE International Conference on Robotics and
  Automation}, 2011, pp. 2520--2525.

\bibitem{aggressive_online_control}
\BIBentryALTinterwordspacing
S.~Li, E.~{\"{O}}zt{\"{u}}rk, C.~D. Wagter, G.~C. H.~E. de~Croon, and D.~Izzo,
  ``Aggressive online control of a quadrotor via deep network representations
  of optimality principles,'' \emph{CoRR}, vol. abs/1912.07067, 2019. [Online].
  Available: \url{http://arxiv.org/abs/1912.07067}
\BIBentrySTDinterwordspacing

\bibitem{robin_thesis}
\BIBentryALTinterwordspacing
R.~Ferede, C.~{De Wagter}, G.~{C. H. E. de Croon}, and D.~Izzo, ``An adaptive
  control strategy for neural network based optimal quadcopter controllers,''
  Master's thesis, TU Delft Aerospace Engineering, 2022. [Online]. Available:
  \url{http://resolver.tudelft.nl/uuid:b43a9703-082c-47c7-a56e-d50794ee8c1c}
\BIBentrySTDinterwordspacing

\bibitem{sanchez_izzo}
C.~Sánchez-Sánchez and D.~Izzo, ``Real-time optimal control via deep neural
  networks: Study on landing problems,'' \emph{Journal of Guidance, Control,
  and Dynamics}, vol.~41, 10 2016.

\bibitem{dario_ekin_earth_venus}
D.~Izzo and E.~Öztürk, ``Real-time guidance for low-thrust transfers using
  deep neural networks,'' \emph{Journal of Guidance, Control, and Dynamics},
  vol.~44, no.~2, pp. 315--327, 2021.

\bibitem{IZZO2022}
\BIBentryALTinterwordspacing
D.~Izzo and S.~Origer, ``Neural representation of a time optimal, constant
  acceleration rendezvous,'' \emph{Acta Astronautica}, 2022. [Online].
  Available:
  \url{https://www.sciencedirect.com/science/article/pii/S0094576522004581}
\BIBentrySTDinterwordspacing

\bibitem{thrust_and_drag_model}
J.~Svacha, K.~Mohta, and V.~R. Kumar, ``Improving quadrotor trajectory tracking
  by compensating for aerodynamic effects,'' \emph{2017 International
  Conference on Unmanned Aircraft Systems (ICUAS)}, pp. 860--866, 2017.

\bibitem{moment_of_inertia}
\BIBentryALTinterwordspacing
S.~Sun, C.~C. de~Visser, and Q.~Chu, ``Quadrotor gray-box model identification
  from high-speed flight data,'' \emph{Journal of Aircraft}, vol.~56, no.~2,
  pp. 645--661, Mar. 2019. [Online]. Available:
  \url{https://doi.org/10.2514/1.c035135}
\BIBentrySTDinterwordspacing

\bibitem{ampl}
R.~Fourer, D.~M. Gay, and B.~W. Kernighan, ``A modeling language for
  mathematical programming,'' \emph{Management Science}, vol.~36, no.~5, pp.
  519--554, 1990.

\bibitem{snopt}
P.~E. Gill, W.~Murray, and M.~A. Saunders, ``Snopt: An sqp algorithm for
  large-scale constrained optimization,'' \emph{SIAM review}, vol.~47, no.~1,
  pp. 99--131, 2005.

\bibitem{kingma2014adam}
D.~P. Kingma and J.~Ba, ``Adam: A method for stochastic optimization,''
  \emph{arXiv preprint arXiv:1412.6980}, 2014.

\bibitem{paparazzi}
B.~Gati, ``Open source autopilot for academic research - the paparazzi
  system,'' in \emph{2013 American Control Conference}, 2013, pp. 1478--1481.

\bibitem{rungekutta}
\BIBentryALTinterwordspacing
J.~Dormand and P.~Prince, ``A family of embedded runge-kutta formulae,''
  \emph{Journal of Computational and Applied Mathematics}, vol.~6, no.~1, pp.
  19--26, 1980. [Online]. Available:
  \url{https://www.sciencedirect.com/science/article/pii/0771050X80900133}
\BIBentrySTDinterwordspacing

\end{thebibliography}

\section{Appendix}

\subsection{Example of more complex track}\label{app:complex_track}

\begin{figure}[htbp]
  \centering
  \includegraphics[width=\columnwidth]{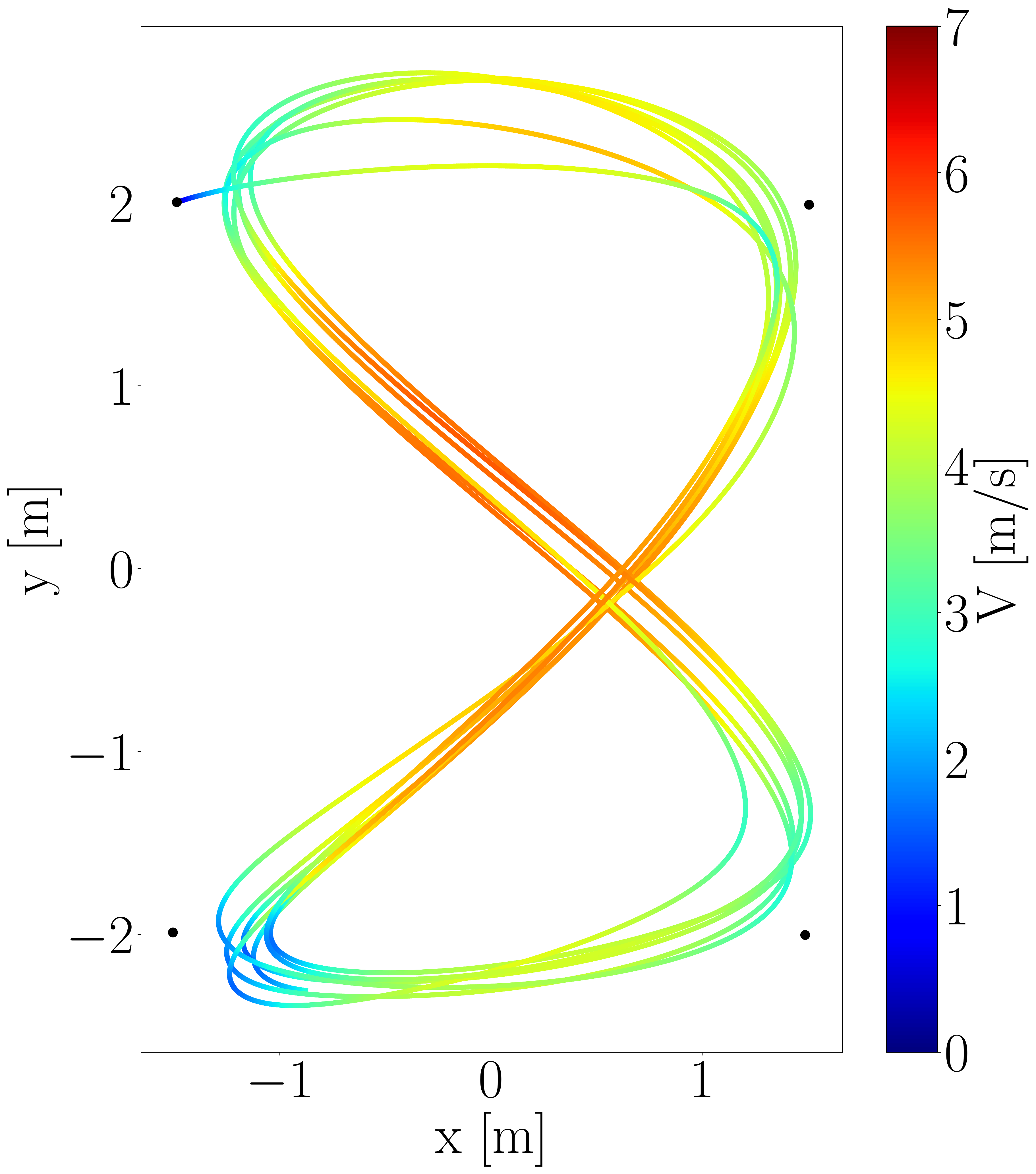}
  \caption{Trajectory (top view) of a flight in simulation with $\epsilon=0.4$. The G\&CNET is trained on 4 sets of sharp turns. During the flight, we inform the G\&CNET of the relative distance and angle of the two upcoming waypoints with two extra inputs.}
  \label{fig:figure_eight}
\end{figure}

Fig.\ref{fig:figure_eight} shows the trajectory of a flight in simulation using a G\&CNET which learned to fly a figure-eight.

\subsection{Effect of battery on maximum RPM limit}\label{app:battery_effect}

\begin{figure}[htbp]
  \centering
  \includegraphics[width=\columnwidth]{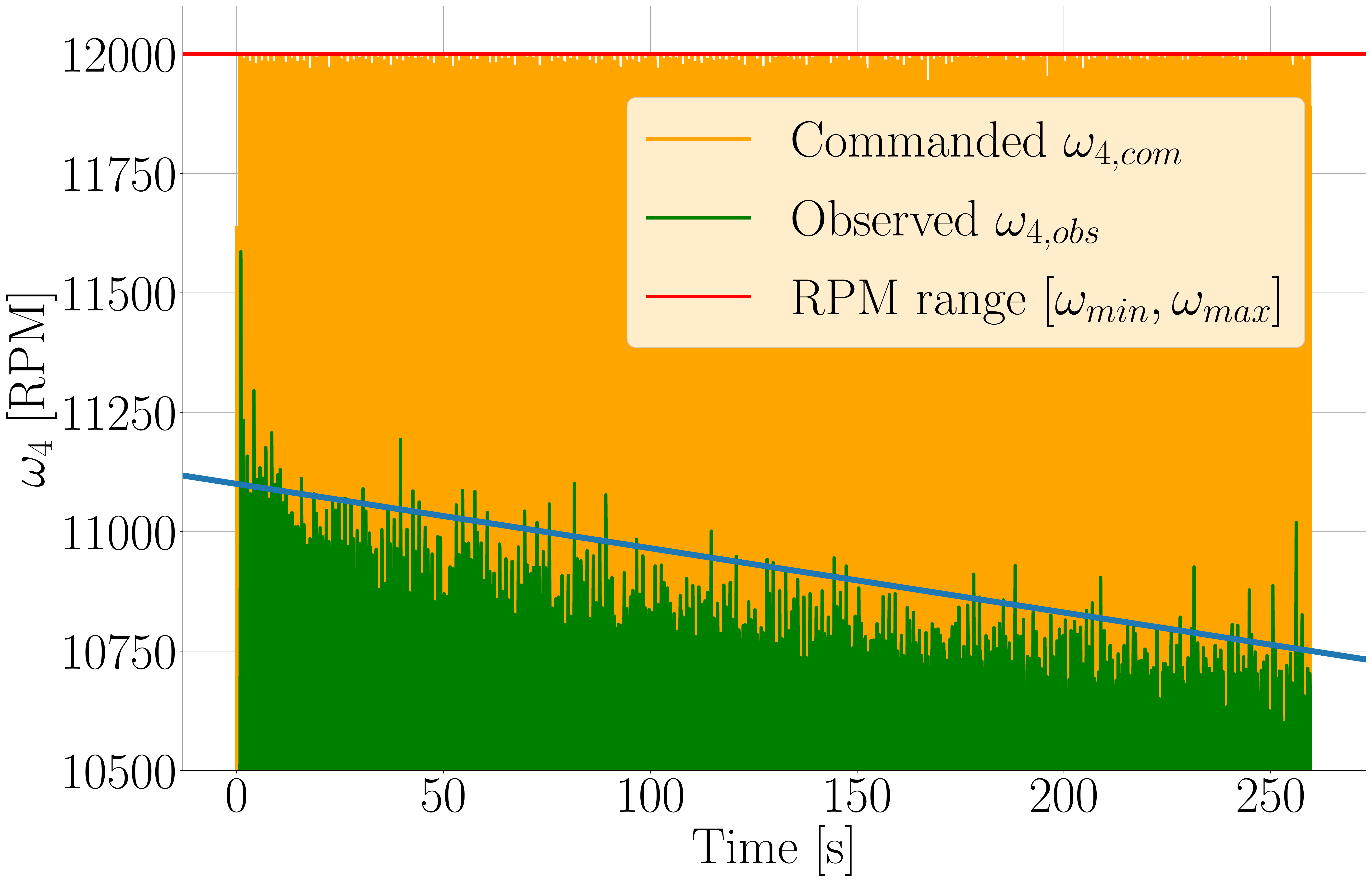}
  \caption{Commanded and observed angular velocity $\omega$ during a test flight with $\epsilon=0.5$ (only rotor number 4 is shown). The blue line roughly indicates the downward trend of $\omega_{max}$.\vspace{-5mm}}
  \label{fig:battery_experiment_rpm_comm_vs_obs}
\end{figure}

We tested a G\&CNET with $\epsilon=0.5$ on the Bebop by flying for $6\si{\min}$ on a $4\times3\si{\meter}$ track. Fig.~\ref{fig:battery_experiment_rpm_comm_vs_obs} is a zoomed-in figure of rotor number 4 (see Fig.~\ref{fig:coordframes}) which saturates most of the time since the flight path mostly consists of right turns. As the battery drains out $\omega_{max}$ decreases by roughly $1\text{ RPM}\si{\per\second}$.

\subsection{Optimal trajectories for G\&CNET training}\label{app:opt_traj_training}

\begin{figure*}[htbp]
  \centering
  \includegraphics[width=\textwidth]{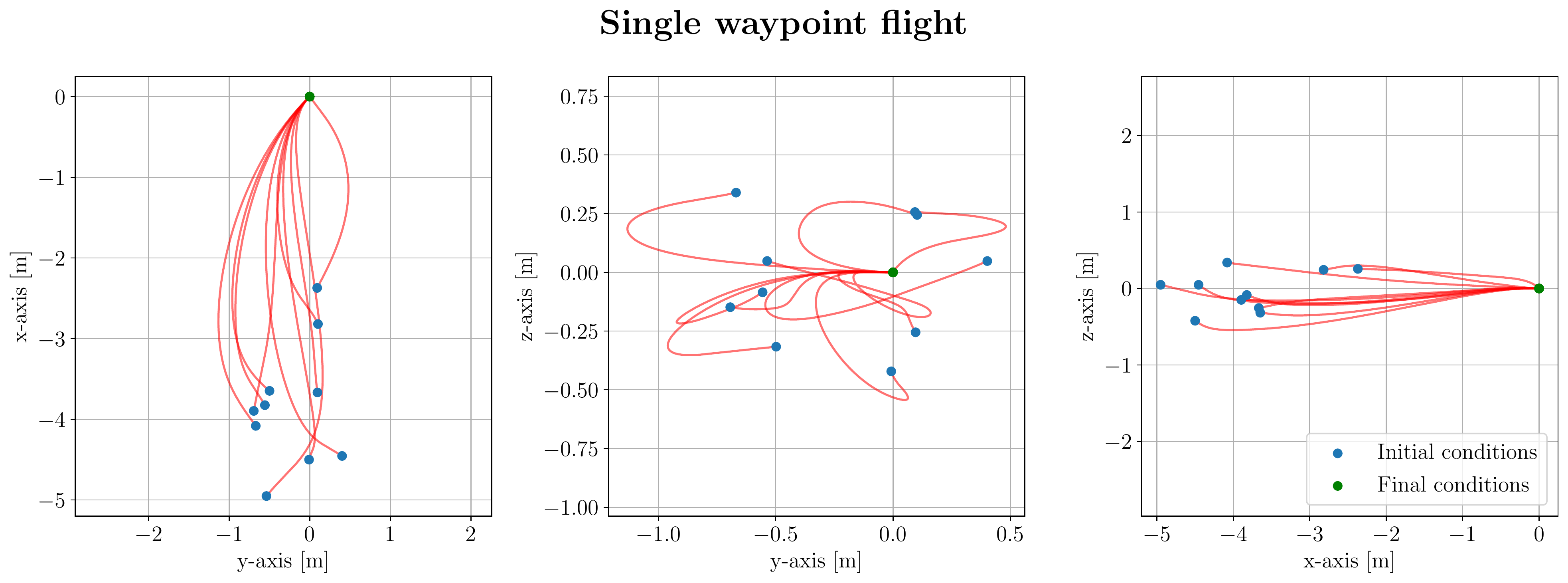}
  \includegraphics[width=\textwidth]{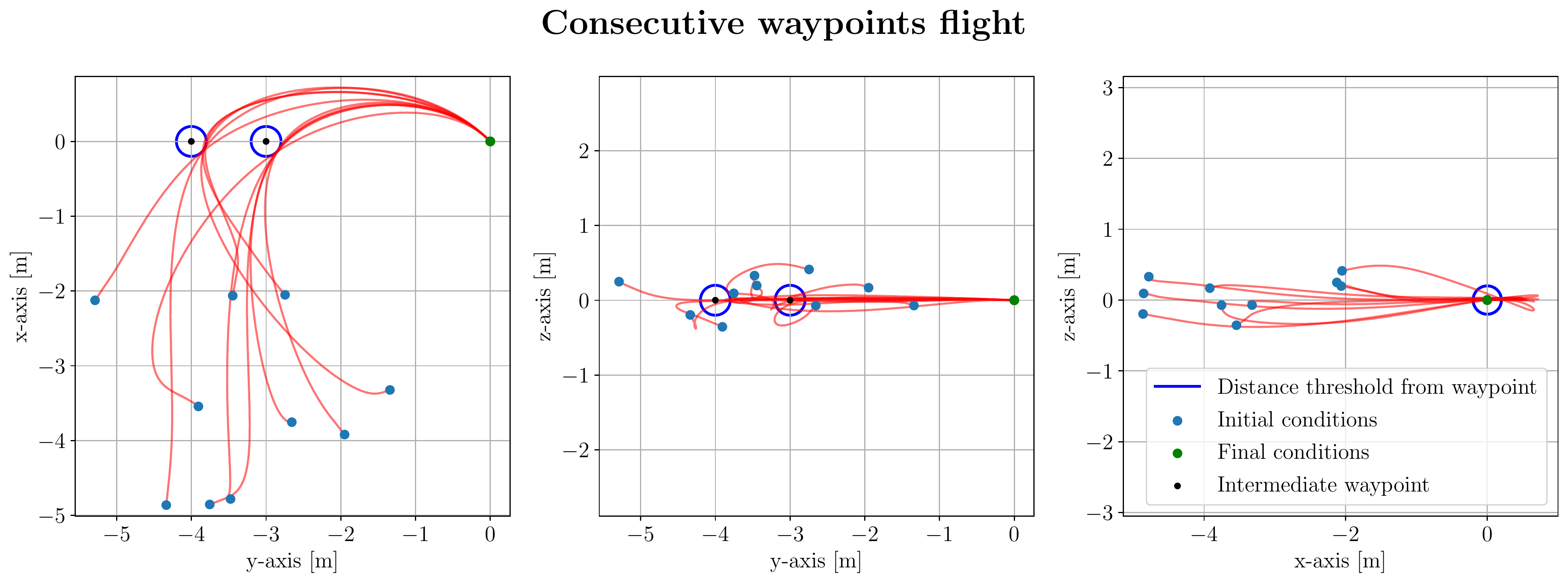}
  \caption{Ten optimal trajectories for single and consecutive waypoints flight as can be found in the training datasets for G\&CNETs.\vspace{-5mm}}
  \label{fig:opt_traj_training}
\end{figure*}

Fig.\ref{fig:opt_traj_training} shows ten optimal trajectories for single and consecutive waypoints flight as can be found in the training datasets for G\&CNETs.

\subsection{Control inputs: DFBC vs G\&CNET}\label{app:control_inputs_dfbc_gcnet}

\begin{figure*}[htbp]
  \centering
  \includegraphics[width=0.95\linewidth]{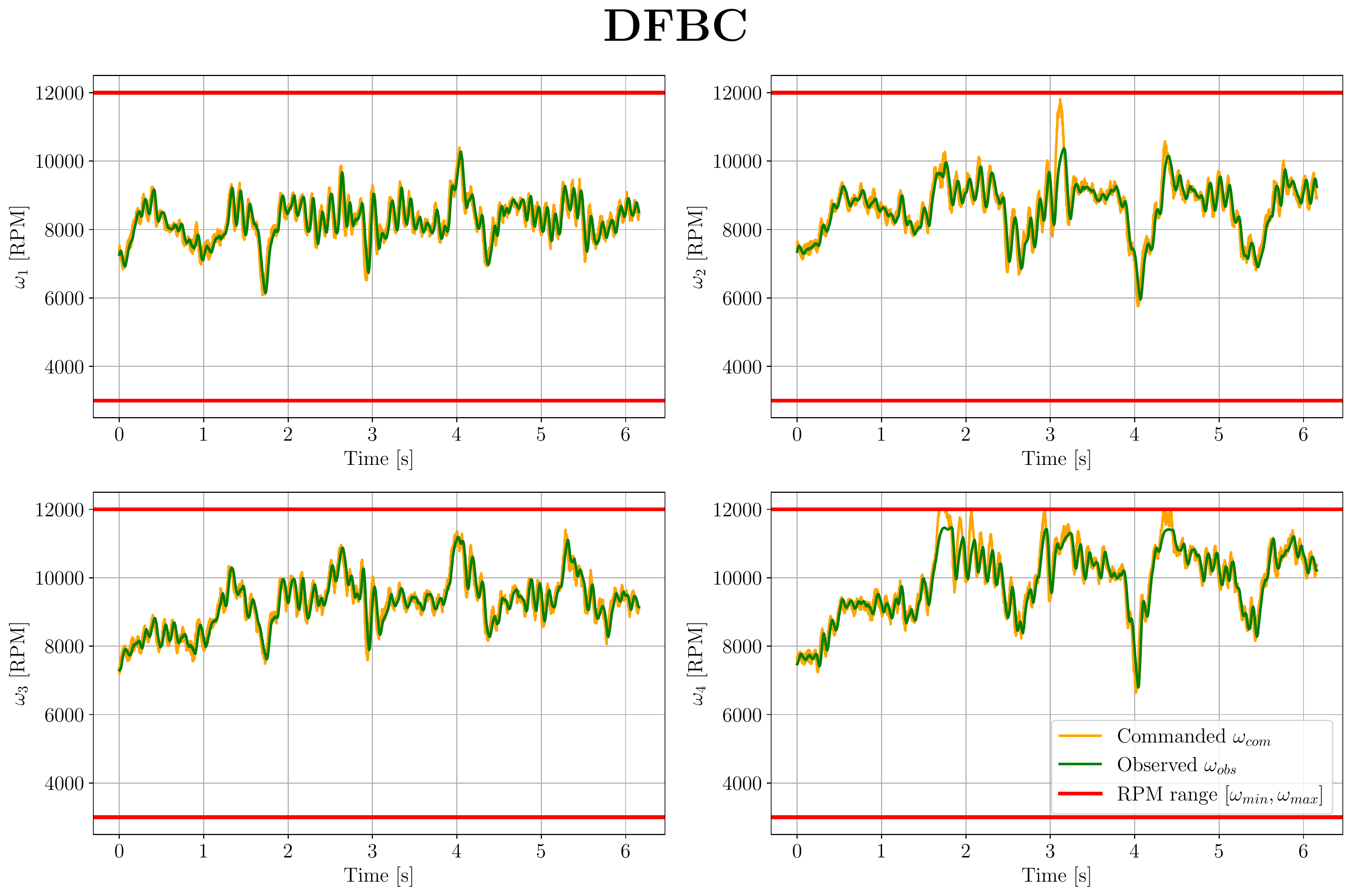}
  \includegraphics[width=0.95\linewidth]{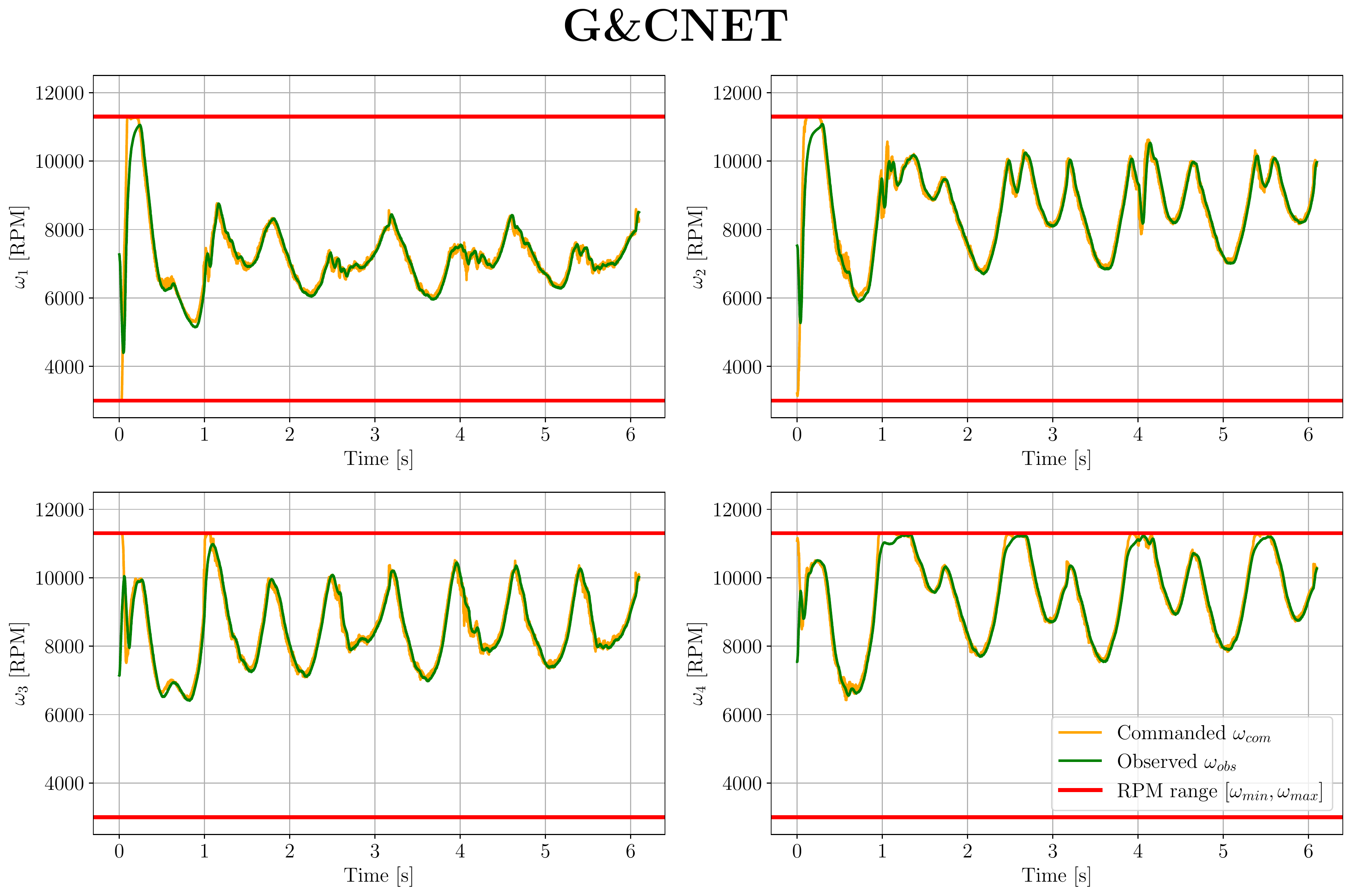}
  \caption{Commanded and observed angular velocities of rotors for DFBC and G\&CNET in Sec\ref{sec:Consecutive WP flight}.\vspace{-5mm}}
  \label{fig:control_inputs_dfbc_gcnet}
\end{figure*}

Fig.\ref{fig:control_inputs_dfbc_gcnet} shows the commanded and observed angular velocities of rotors for the real flights using a Differential-Flatness-Based-Controller (DFBC) and a Guidance \& Control Network (see Sec.\ref{sec:Consecutive WP flight}).

\end{document}